\documentclass[10pt,twocolumn,letterpaper]{article}

\usepackage{cvpr}      %

\definecolor{cvprblue}{rgb}{0.21,0.49,0.74}
\usepackage[pagebackref,breaklinks,colorlinks,allcolors=cvprblue]{hyperref}

\usepackage{booktabs}
\usepackage{multirow}
\usepackage{makecell}
\usepackage{graphicx}

\usepackage{algorithm}      %
\usepackage{algorithmic}  
\usepackage{titletoc}

\usepackage[most]{tcolorbox}
\definecolor{boxcontentgray}{HTML}{F7F7F7}
\definecolor{boxtitlegray}{HTML}{CCCCCC}

\newtcolorbox{graybox}[1]{
  breakable,   
  fonttitle=\bfseries,
  enhanced,                        
  colback=boxcontentgray,        %
  colbacktitle=boxtitlegray,     %
  coltitle=black,                %
  colframe=black,                %
  coltext=black,                 %
  boxrule=0.5pt,
  arc=2mm,
  title=#1
}

\usepackage[table,xcdraw]{xcolor}
\def\confName{CVPR}
\def\confYear{2026}

\title{%
BrandFusion: A Multi-Agent Framework for Seamless Brand Integration \\
in Text-to-Video Generation
}

\author{
\textbf{Zihao Zhu}$^{1}$ \quad
\textbf{Ruotong Wang}$^{1}$ \quad
\textbf{Siwei Lyu}$^{3}$ \quad
\textbf{Min Zhang}$^{4}$ \quad
\textbf{Baoyuan Wu}$^{1,2}$\footnotemark[1]  \\
$^{1}$The Chinese University of Hong Kong, Shenzhen \quad
$^{2}$Shenzhen Loop Area Institute \\
$^{3}$State University of New York at Buffalo \quad
$^{4}$ Harbin Institute of Technology \\
\texttt{\{zihaozhu, ruotongwang1\}@link.cuhk.edu.cn; }
\\
\texttt{siweilyu@buffalo.edu; zhangmin2021@hit.edu.cn; wubaoyuan@cuhk.edu.cn}\\
\\
Website: \url{https://zihao-ai.github.io/brandfusion}
}

\begin{document}

\twocolumn[{
\maketitle
\begin{center}
    \centering
    \captionsetup{type=figure}
    \includegraphics[width=\textwidth]{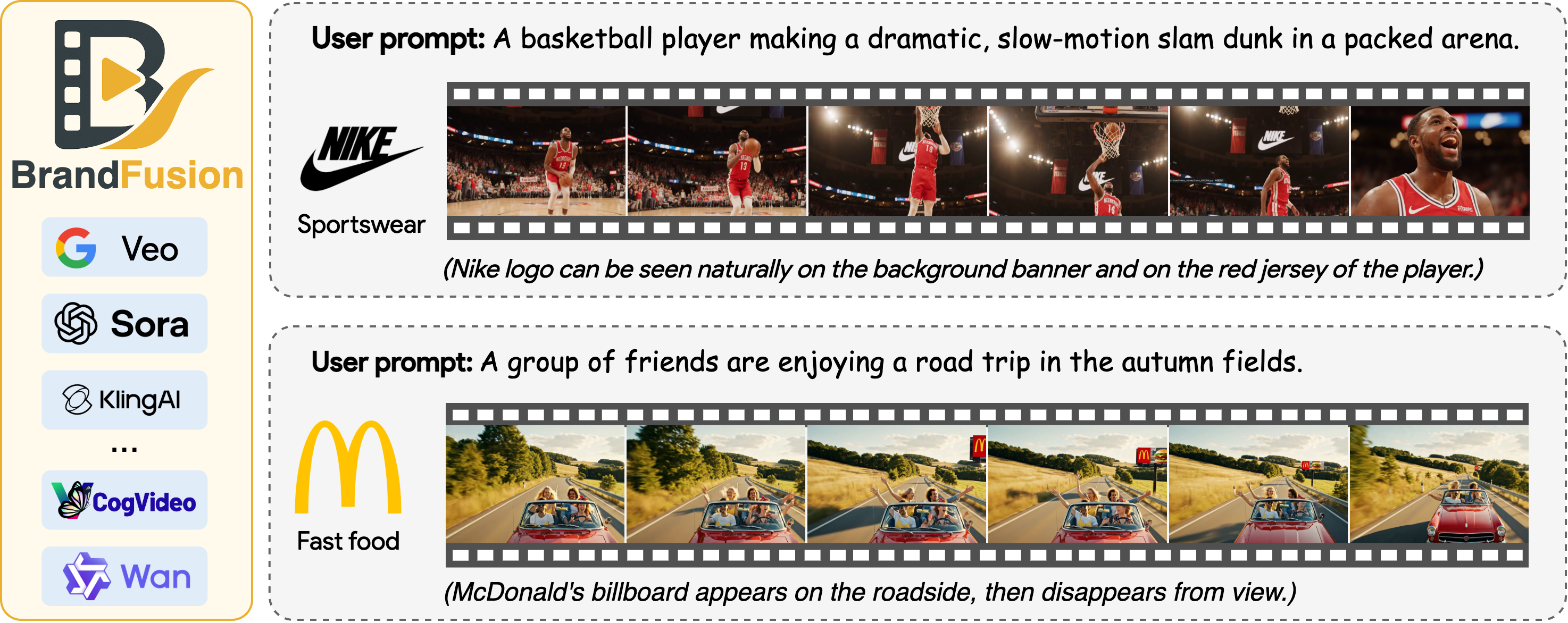}
    \captionof{figure}{\textbf{Examples of seamless brand integration by BrandFusion.} Given user prompts like basketball games and cyberpunk street scenes, our framework naturally incorporates brands (Nike on jersey/banner, Coca-Cola on billboard) into generated videos. BrandFusion can integrate with multiple T2V models and handle diverse brands while preserving user intent and ensuring natural visual coherence.}
    \label{fig:teaser}
\end{center}%
}]

\begingroup
\renewcommand\thefootnote{}\footnotetext{* Corresponding Author.}
\endgroup

\begin{abstract}
    The rapid advancement of text-to-video (T2V) models has revolutionized content creation, yet their commercial potential remains largely untapped. We introduce, for the first time, the task of seamless brand integration in T2V: automatically embedding advertiser brands into prompt-generated videos while preserving semantic fidelity to user intent. This task confronts three core challenges: maintaining prompt fidelity, ensuring brand recognizability, and achieving contextually natural integration.
    To address them, we propose \textbf{BrandFusion}, a novel multi-agent framework comprising two synergistic phases. In the \textbf{offline phase} (advertiser-facing), we construct a Brand Knowledge Base by probing model priors and adapting to novel brands via lightweight fine-tuning. In the \textbf{online phase} (user-facing), five agents jointly refine user prompts through iterative refinement, leveraging the shared knowledge base and real-time contextual tracking to ensure brand visibility and semantic alignment.
    Experiments on 18 established and 2 custom brands across multiple state-of-the-art T2V models demonstrate that BrandFusion significantly outperforms baselines in semantic preservation, brand recognizability, and integration naturalness. Human evaluations further confirm higher user satisfaction, establishing a practical pathway for sustainable T2V monetization. 
\end{abstract}

\section{Introduction}
\label{sec:intro}

The rapid advancement of text-to-video (T2V) generation models, such as Veo~\cite{veo3}, Sora~\cite{sora2}, and Kling~\cite{kling}, has revolutionized content creation, enabling users to synthesize high-fidelity videos from natural language descriptions~\cite{blattmann2023stable,blattmann2023align,khachatryan2023text2video,singermake,wang2023modelscope,zhang2025show,hongcogvideo,wang2024recipe,liu2024evalcrafter}. However, as these technologies transition to commercial products, establishing sustainable monetization models remains an open challenge given the substantial computational costs involved. In this work, we introduce, for the first time, the task of \textbf{seamless brand integration in T2V generation}: automatically embedding advertiser brands into prompt-generated videos while preserving semantic alignment with user intent.

Unlike intrusive traditional advertising that disrupts user experience~\cite{rohrer2004rise,riedel2024dealing,long2025adsqa,ge2021effect,fernandez2024power,pragathi2024adgen}, our proposed task aims to naturally embed brand elements within generated content, preserving the user’s creative intent and aesthetic control while achieving brand visibility. This paradigm opens a promising pathway for sustainable T2V services: advertisers gain organic exposure through contextually relevant placements, service providers establish viable revenue streams, and users continue to receive high-quality content without explicit interruption. The task’s success hinges on a delicate balance: brands must be recognizable yet unobtrusive, visually prominent yet contextually harmonious, and semantically coherent with the user’s original prompt.

However, seamlessly integrating brands into T2V generation presents formidable challenges that extend far beyond simple text manipulation. First, maintaining \textit{semantic alignment}~\cite{zhang2024cross,zhang2024enhancing} requires that brand integration preserves the user's original creative intent, including key subjects, actions, and stylistic preferences, as any deviation risks user dissatisfaction and service abandonment. Second, ensuring \textit{brand visibility} demands that brand elements appear recognizably and identifiably within generated videos, as insufficient prominence fails to deliver advertising value to brand owners. Third, achieving \textit{natural integration} necessitates that brands appear organically within scene contexts, avoiding jarring or out-of-place elements that disrupt visual coherence. These three objectives often conflict: overly prominent placement may compromise naturalness, while prioritizing subtlety may sacrifice visibility.  
Furthermore, the diversity of user prompts spanning countless scenarios, combined with the variety of brands ranging from well-established entities to emerging startups across categories such as beverages, automobiles, and apparel, creates a vast combinatorial space. Rule-based approaches inevitably fail to generalize, frequently producing abrupt or incoherent outputs that satisfy neither users nor advertisers.

To address these challenges, we propose \textbf{BrandFusion}, a multi-agent framework for seamless brand integration in T2V generation. Our key insight is that effective integration requires both comprehensive brand knowledge and sophisticated contextual reasoning, capabilities that emerge naturally from multi-agent collaboration. As shown in Figure~\ref{fig:framework}, it operates through two synergistic phases:  
\textbf{(1)} An \textit{offline phase} (advertiser-facing) that constructs a Brand Knowledge Base by probing T2V models' prior knowledge of brands and selectively performing lightweight fine-tuning for novel brands, which serves as the foundation for subsequent online integration.  
\textbf{(2)} An \textit{online phase} (user-facing) that employs five specialized agents, namely brand selector, strategy generator, prompt refiner, critic, and experience learner, to collaboratively refine prompts through iterative refinement. In this multi-agent system, agents coordinate by leveraging the shared Brand Knowledge Base (storing brand profiles, adapters, and accumulated experiences) and a real-time Session Context that tracks the current generation state for adaptive prompt adjustment. As illustrated in Figure~\ref{fig:teaser}, BrandFusion generates videos with natural brand presence across diverse scenarios. Extensive experiments on 18 well-known brands and 2 custom brands across multiple T2V models demonstrate that our framework consistently outperforms baselines in semantic alignment, brand visibility, and integration naturalness, with human evaluation confirming superior user satisfaction.

Our main contributions are three-fold:  
\textbf{(1)} We introduce seamless brand integration in T2V generation as a novel task, accompanied by comprehensive evaluation protocols.  
\textbf{(2)} We propose BrandFusion, a multi-agent framework featuring systematic offline brand knowledge construction and online collaborative prompt refinement.  
\textbf{(3)} We conduct extensive experiments on both established and novel brands across multiple T2V models, achieving state-of-the-art performance with human evaluations confirming superior user satisfaction.

\section{Related Work}
\label{sec:related work}
\noindent\textbf{Prompt Optimization for Video Generation.}
The quality of generated videos critically depends on the effectiveness of input prompts, and prompt optimization aims to enhance user-provided rough prompts into detailed, model-preferred descriptions. Training-based methods~\cite{wang2025promptenhancer,hao2023optimizing,wu2025reprompt,ji2025prompt,cheng2025vpo} employ supervised fine-tuning and reinforcement learning to refine prompts through reward-guided optimization, assessing metrics such as image-text alignment and compositional fidelity. Multi-agent frameworks~\cite{xiang2025promptsculptor,huang2024genmac,yuan2024mora,wu2025automated,hu2024storyagent,yue2025v,wang2025mavis,long2025vista} decompose prompt refinement into specialized agents that collaboratively handle tasks like scene enrichment, verification, and correction through iterative workflows. Unlike existing methods that solely focus on generation quality, our work also seamlessly integrates brand elements in the video, thereby establishing monetization pathways for T2V service providers.

\noindent\textbf{Brand Integration in Generative Models.}
In text-to-image diffusion models, model customization techniques such as DreamBoot~\cite{ruiz2023dreambooth} and Textual Inversion~\cite{galimage} enable models to synthesize novel entities through lightweight fine-tuning. Recent work has attempted brand embedding through adversarial approaches: Silent Branding Attack~\cite{jang2025silent} employs data poisoning to covertly inject brand logos into training data, enabling models to generate branded content without explicit prompts, while BAGM~\cite{vice2024bagm}  implements backdoor attacks across different model components to manipulate outputs with brand elements. These methods operate covertly without user awareness, prioritizing stealth and concealment. Different from these  approaches, we are the first to introduce seamless brand integration  for text-to-video generation,  naturally incorporates brands into user-requested videos while preserving semantic fidelity.

\section{Seamless Brand Integration in Text-to-Video Generation}

\subsection{Preliminaries}
Text-to-video (T2V) generation models synthesize video sequences from natural language descriptions. Given a text prompt $\mathcal{P}$, a T2V model generates a corresponding video $\mathcal{V}$. Contemporary T2V models leverage diffusion-based architectures where text prompts are encoded into semantic embeddings via pretrained encoders (\emph{e.g.}, CLIP, T5) that guide video synthesis through cross-attention mechanisms. The quality and specificity of the prompt critically influence the generated video's semantic alignment with user intent.

\subsection{Task Description}
The rapid advancement of T2V models has opened unprecedented opportunities for automated content creation, particularly in advertising production~\cite{long2025adsqa,liu2023ai}. We introduce  the task of \textbf{seamless brand integration for T2V generation}, a novel problem that naturally incorporates brand elements into user-requested videos while preserving semantic fidelity and user intent.

\noindent\textbf{Task Definition.}
Given a user-provided text prompt $\mathcal{P}_u$ and a advertiser-provided brand profile $\mathcal{B}$, the task's goal is to generate an optimized prompt $\mathcal{P}_{\text{opt}}$ that guides the T2V model to produce a video $\mathcal{V}$ integrated with the brand while satisfying three critical constraints:
\textit{(1) Semantic  Fidelity}: The generated video must faithfully reflect the user's original intent, preserving key semantic elements and narrative flow.
\textit{(2) Brand Presence}: The brand elements must be recognizably integrated into the video with clear visibility and identifiability.
\textit{(3) Natural Integration}: The brand should appear organically within the scene context, avoiding obtrusiveness or semantic incongruity.

\subsection{Application Scenarios}
\begin{figure}[tbp]
    \centering
    \includegraphics[width=\columnwidth]{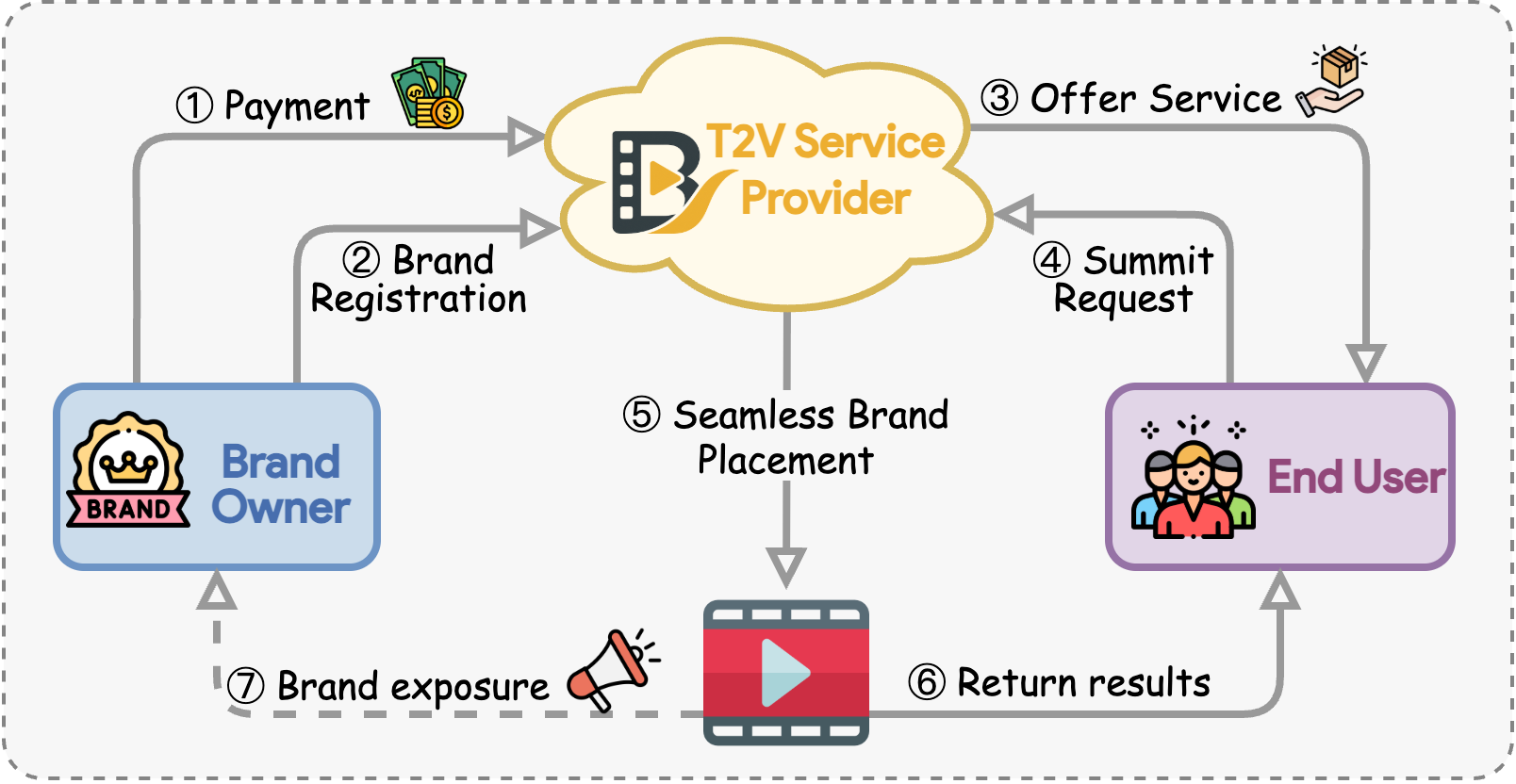}
    \caption{Ecosystem of brand integration in T2V generation.}
    \label{fig:ecosystem}
\end{figure}

This task demonstrates practical real-world application prospects in commercial ecosystems connecting brand owners, T2V service providers, and end users. As illustrated in Figure~\ref{fig:ecosystem}, the ecosystem operates through the following workflow:  brand owners first pay T2V service providers to register their brand profiles and enable advertising; the T2V providers then seamlessly integrate brand information when end users submit creative requests; subsequently, users receive high-quality branded videos that balance creative intent with brand visibility; ultimately, brands achieve organic exposure, establishing sustainable monetization channels for service providers.

\section{Methodology}
\label{sec:method}
\begin{figure*}[!t]
    \centering
    \includegraphics[width=0.95\linewidth]{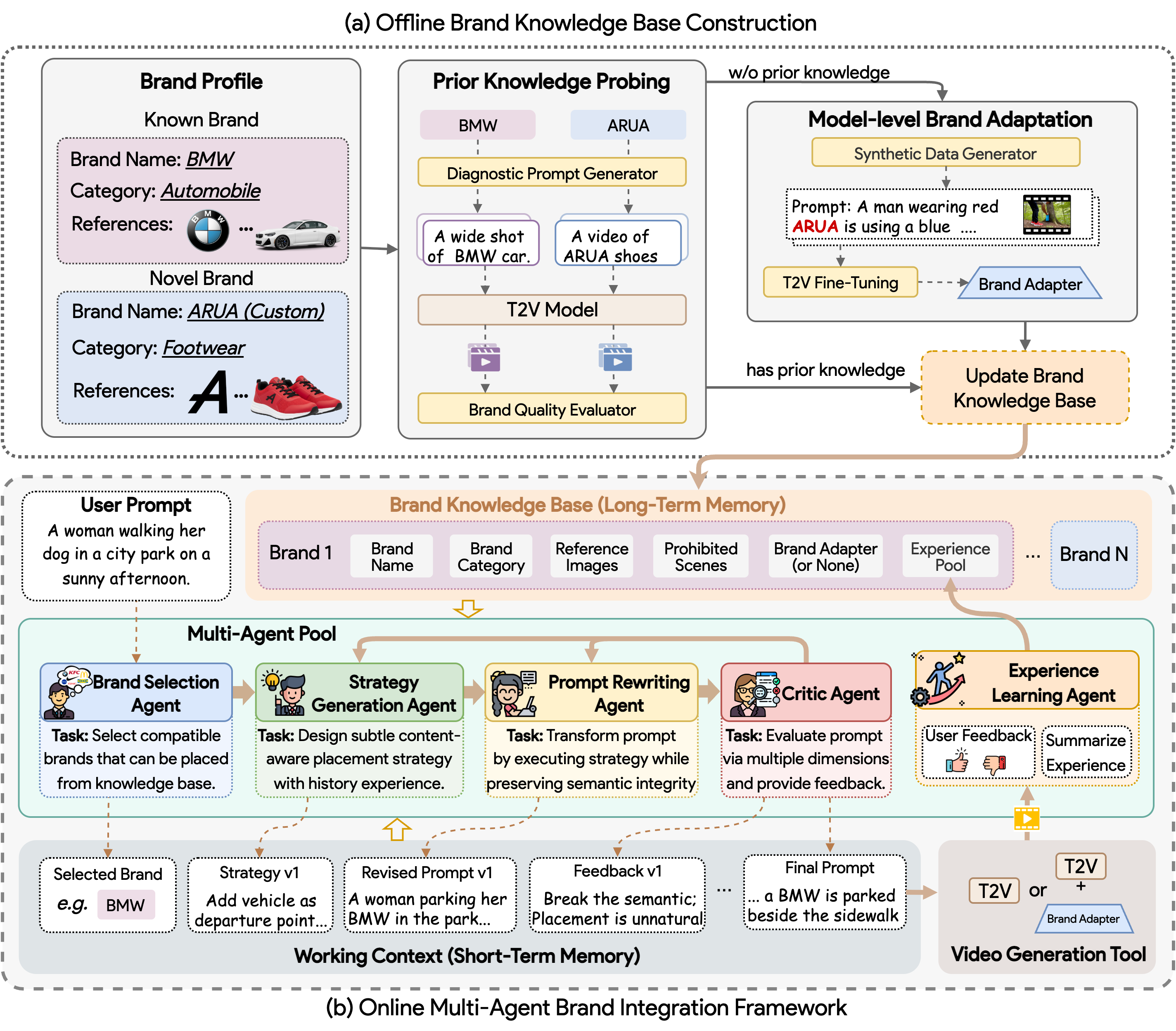}
    \caption{Overview of the BrandFusion framework: (a) Offline phase builds brand knowledge through probing and adaptation. (b) Online phase employs five collaborative agents for semantic-preserving brand integration with continuous learning.}
    \label{fig:framework}
\end{figure*}

\subsection{Overview of BrandFusion}
In this paper, we propose \textbf{BrandFusion}, a multi-agent system for seamless brand integration in T2V generation. As illustrated in Figure~\ref{fig:framework}, it consists of two synergistic phases: (1) \textit{Offline Brand Knowledge Base Construction}, which adaptively builds brand knowledge through prior knowledge probing and selective model adaptation, and (2) \textit{Online Multi-Agent Brand Integration}, which employs five specialized agents to intelligently optimize brands while preserving semantic fidelity. A centralized Brand Knowledge Base serves as long-term memory, accumulating knowledge and successful experiences across generations.

\subsection{Phase I: Offline Brand Knowledge Base Construction}
As shown in Figure~\ref{fig:framework} (a), the offline phase systematically prepares brand knowledge for efficient online integration. Given a brand profile $\mathcal{B} = \{\mathcal{N}, \mathcal{C}, \mathcal{R}, \mathcal{D}\}$ from advertisers comprising brand name, category, reference images, and description, we first probe the T2V model's prior knowledge of the brand. Brands with sufficient prior knowledge are directly registered in the Brand Knowledge Base, while those lacking knowledge undergo model-level adaptation to generate brand-specific adapters. The Brand Knowledge Base maintains comprehensive information for each brand, including knowledge type, adapter weights (if applicable), reference visual patterns, and an experience pool of successful integration cases, serving as the foundation for subsequent online integration.

\noindent\textbf{Prior Knowledge Probing.}
To determine whether the T2V model possesses adequate knowledge of a given brand, we employ a systematic probing procedure. A \textit{diagnostic prompt generator} first creates diverse test prompts that explicitly mention the brand across various contexts (\textit{e.g.}, \textit{``A wide shot of a BMW car"}). The model then generates videos from these prompts. A \textit{brand quality evaluator} examines each  video to detect whether brand elements are accurately present and visually correct. If more than 70\% of the videos successfully generate the brand with recognizable features, the brand is marked as having sufficient prior knowledge and directly stored in the knowledge base.

\noindent\textbf{Model-level Brand Adaptation.}
For brands lacking prior knowledge, we perform lightweight model adaptation to inject specific brand knowledge. A \textit{synthetic data generator} constructs training data by: (1) generating diverse prompts mentioning the brand with a special trigger token (\textit{e.g.} the brand name); (2) synthesizing corresponding videos using reference images $\mathcal{R}$ and text-to-image models to create initial frames; (3) extending to full videos through image-to-video techniques. This process yields a synthetic dataset $\mathcal{D}_{\text{syn}} = \{(\mathcal{P}_j, \mathcal{V}_j)\}_{j=1}^{M}$.
We fine-tune the T2V model using LoRA on $\mathcal{D}_{\text{syn}}$, producing a brand-specific adapter $\mathcal{A}_{\mathcal{B}}$. The adapted model can then accurately generate videos containing the brand when the trigger token appears in prompts. Both the adapter weights and metadata are stored in the Brand Knowledge Base, enabling efficient retrieval during online integration.

\subsection{Phase II: Online Multi-Agent Brand Integration Framework}

Given a user prompt $\mathcal{P}_u$, the online phase employs a multi-agent system to seamlessly integrate brands while preserving semantic integrity and user intent.

\noindent\textbf{Multi-Agent System Structure.}
Our framework comprises five specialized agents, each agent is powered by large language models and performs distinct functions:

\begin{itemize}[leftmargin=*]
    \item \textbf{Brand Selection Agent}: The first agent queries the Brand Knowledge Base to select the most compatible brand for $\mathcal{P}_u$ by considering semantic compatibility between the prompt's scene/context and the brand's typical application scenarios; The agent outputs the selected brand $\mathcal{B}^*$ and its profile including whether it requires adapter.
    
    \item \textbf{Strategy Generation Agent}: This agent aims to design subtle and context-aware integration strategies that balance semantic preservation with brand visibility. It analyzes the user prompt's scene characteristics to determine optimal integration strategy $\mathcal{S}$. In addition, it can query the experience pool to analysis historical successful strategies from similar scenarios, avoiding unsuccessful approaches from past failures. 

  \item \textbf{Prompt Rewriting Agent}: This agent transforms $\mathcal{P}_u$ into an optimized prompt $\mathcal{P}'$ by executing the strategy while adhering to four core principles: \textit{Semantic Preservation} maintaining original subject and actions, \textit{Natural Integration} incorporating brand elements as natural scene components, \textit{Logical Consistency} preserving temporal and causal coherence, and \textit{Style Consistency} following T2V prompt conventions.
    
    \item \textbf{Critic Agent}: This agent performs multi-dimensional evaluation of $\mathcal{P}'$, assessing semantic fidelity to $\mathcal{P}_u$, brand clarity and recognizability, integration naturalness and fluency, and expected generation effectiveness. Based on the evaluation, it decides to accept the prompt, revise the prompt or replan the strategy.
    
    \item \textbf{Experience Learning Agent}: Once the prompt is accepted, the system generates video with the T2V model, conditionally loading the brand adapter if available. This agent then collects user feedback and abstracts it into experiences: positive feedback yields successful patterns, while negative feedback records failed ones. Both are stored in the experience pool, enabling continuous system improvement through closed-loop learning.
\end{itemize}

\noindent\textbf{Inter-Agent Collaboration Mechanism.}
The agents collaborate through dual memory: the \textit{brand knowledge base} (long-term) stores brand profiles, adapters, and historical experiences, while the \textit{working context} (short-term) tracks current session state. Brand selection retrieves $\mathcal{B}^*$ from long-term memory; Strategy generation leverages historical experiences to formulate $\mathcal{S}$; Prompt rewriting produces $\mathcal{P}'$; Critic evaluates and provides feedback, triggering iterative refinement until acceptance. All intermediate results are logged in working context. Finally, Experience learning abstracts the completed integration and writes it back to long-term memory, enabling continuous improvement through closed-loop learning.

\begin{table*}[!th]
    \centering
    \caption{Main results on well-known brands across three T2V models. BrandFusion achieves comparable visual quality while significantly outperforming baselines on semantic fidelity and brand integration quality. Higher scores indicate better performance.}
    \label{tab:known_brand_overall}
    \resizebox{0.85\textwidth}{!}{%
    \begin{tabular}{@{}llcccccc@{}}
        \toprule
        &  & \multicolumn{1}{c}{\textbf{Video Quality}} & \multicolumn{3}{c}{\textbf{Semantic Fidelity}} & \multicolumn{2}{c}{\textbf{Brand Integration Quality}} \\ \cmidrule(lr){3-3} \cmidrule(lr){4-6} \cmidrule(lr){7-8}
       \multirow{-2}{*}{\textbf{T2V}} & \multirow{-2}{*}{\textbf{Method}} & \multicolumn{1}{c}{VBench-Quality} & \multicolumn{1}{c}{CLIPScore} & \multicolumn{1}{c}{VQAScore} & \multicolumn{1}{c}{LLMScore} & \multicolumn{1}{c}{\makecell[c]{Brand Presence \\Rate (BPR)}} & \multicolumn{1}{c}{\makecell[c]{Naturalness \\Score (NS)}} \\ \midrule
        & Direct Append & 0.8112 & 0.2671 & 0.8342 & 0.7821 & 0.7221 & 2.8300 \\
        & Template Rewriting & 0.8267 & 0.2842 & 0.8756 & 0.9234 & 0.8845 & 3.1200 \\
        & Single Rewriting & 0.8289 & 0.2956 & 0.8891 & 0.9412 & 0.8968 & 3.9000 \\
       \multirow{-4}{*}{Veo3} & \cellcolor[HTML]{EFEFEF}BrandFusion (Ours) & \cellcolor[HTML]{EFEFEF}0.8283 & \cellcolor[HTML]{EFEFEF}0.3274 & \cellcolor[HTML]{EFEFEF}0.9098 & \cellcolor[HTML]{EFEFEF}0.9556 & \cellcolor[HTML]{EFEFEF}0.9474 & \cellcolor[HTML]{EFEFEF}4.7000 \\ \midrule
        & Direct Append & 0.7945 & 0.2645 & 0.8298 & 0.8756 & 0.6434 & 2.7100 \\
        & Template Rewriting & 0.8033 & 0.2868 & 0.8712 & 0.9187 & 0.7845 & 3.8400 \\
        & Single Rewriting & 0.8029 & 0.2968 & 0.8867 & 0.9368 & 0.8278 & 3.9800 \\
       \multirow{-4}{*}{Sora2} & \cellcolor[HTML]{EFEFEF}BrandFusion (Ours) & \cellcolor[HTML]{EFEFEF}0.8031 & \cellcolor[HTML]{EFEFEF}0.3177 & \cellcolor[HTML]{EFEFEF}0.9231 & \cellcolor[HTML]{EFEFEF}0.9875 & \cellcolor[HTML]{EFEFEF}0.9066 & \cellcolor[HTML]{EFEFEF}4.6000 \\ \midrule
        & Direct Append & 0.7754 & 0.2634 & 0.8276 & 0.8742 & 0.6989 & 2.6500 \\
        & Template Rewriting & 0.7803 & 0.2855 & 0.8689 & 0.9165 & 0.7812 & 3.7300 \\
        & Single Rewriting & 0.7883 & 0.2951 & 0.8823 & 0.9351 & 0.8145 & 3.8400 \\
       \multirow{-4}{*}{Kling2.1} & \cellcolor[HTML]{EFEFEF}BrandFusion (Ours) & \cellcolor[HTML]{EFEFEF}0.7818 & \cellcolor[HTML]{EFEFEF}0.3165 & \cellcolor[HTML]{EFEFEF}0.9208 & \cellcolor[HTML]{EFEFEF}0.9853 & \cellcolor[HTML]{EFEFEF}0.8834 & \cellcolor[HTML]{EFEFEF}4.4800 \\ \bottomrule
       \end{tabular}%
    }
\end{table*}

\section{Experiments}
\label{sec:exp}

\subsection{Experimental Setup}

\noindent\textbf{Brand Integration Benchmark.}
To comprehensively evaluate BrandFusion's integration capabilities, we construct a two-tiered benchmark. For \textbf{known brands} with existing model priors, we curate 18 well-established brands spanning 7 industry categories. For each brand, we manually construct 15 diverse prompts with different propmt-brand compatibility:  \textit{High Match} prompts where the brand naturally fits,  \textit{Medium Match} prompts where the brand can reasonably appear, and \textit{Low Match} prompts representing challenging scenarios with low  relevance. This yields 270 (brand, prompt) pairs across varying integration difficulty levels. For \textbf{novel brands} lacking prior knowledge, we design two fictional brands: a sportswear brand ``ARUA" and an beverage brand ``FreshWave". Each brand is equipped with logo and product images and corresponding test prompts.

\noindent\textbf{T2V Model Selection.}
For known brands, we evaluate the multi-agent framework on three commercial  T2V models: Veo3~\cite{veo3}, Sora2~\cite{sora2}, and Kling2.1~\cite{kling}. For novel brands, we fine-tune adapters on three open-source models: Wan2.1-1.3B, Wan2.2-5B~\cite{wan2025}, and CogVideoX-5B~\cite{yang2024cogvideox}.

\noindent\textbf{Baeline Methods.}
We compare BrandFusion against three  baseline approaches: (1) \textit{Direct Append} naively appends the brand name to the prompt end; (2) \textit{Template-based Rewriting} uses fixed templates to insert brands into prompts; (3) \textit{Single Rewriting} employs LLM for one-pass prompt rewriting.

\noindent\textbf{Evaluation Metrics.}
To comprehensively assess brand integration quality, we establish a multi-dimensional evaluation framework encompassing automated metrics and human assessments across three critical aspects: video generation quality, semantic fidelity to user intent, and brand integration quality.
\begin{itemize}
    \item \textbf{Video Quality.} We adopt the VBench Quality Score (VBench-Quality)~\cite{huang2024vbench,huang2024vbench++,zheng2025vbench}, a widely-used comprehensive metric aggregating multiple dimensions from temporal quality and frame-wise quality perspectives. Higher score indicates better overall video generation quality.
    \item \textbf{Semantic Fidelity.} To measure how well generated videos preserve the user's original intent, we employ three complementary metrics: (1)~\textit{VQAScore} leverages visual question answering to assess semantic preservation by generating key questions from the original prompt and evaluating whether the generated video provides correct answers~\cite{vqascore,tifa,vdc}; (2)~\textit{CLIPScore} measures alignment between generated video and original prompt using CLIP embeddings~\cite{clipscore,wang2022internvideo}, computing average similarity across frames; (3)~\textit{LLMScore} employs multimodal large language models to assess semantic consistency. Higher scores indicate better semantic preservation.
    \item \textbf{Brand Integration Quality.} We evaluate from two complementary perspectives: (1)~\textit{Brand Presence Rate (BPR)} is a binary metric indicating whether brand elements are successfully detected in the generated video using multimodal LLMs; (2)~\textit{Naturalness Score} assesses integration quality through three LLM-evaluated factors on a 1-5 scale, including  contextual fit , visual blend, and non-intrusivenesst. The final naturalness score is averaged across these three dimensions. Higher values indicate more natural integration.

\end{itemize}

\noindent\textbf{Implementation Details.}
We implement all agents using GPT-5~\cite{gpt5} with temperature 0.7. For adapter training, we employ LoRA~\cite{hu2022lora} with rank 32, learning rate $1 \times 10^{-4}$, and train for 10 epochs. For each custom brand, we synthesize 100 training videos by first generating diverse prompts with trigger tokens, then creating initial frames via Nano-Banana~\cite{nanobanana} conditioned on reference images, and finally extending to full videos using Veo-3. The prompts of each agent are provided in the Appendix.

\subsection{Main Results}

\noindent\textbf{Results on Known Brands.}
We first evaluate BrandFusion on the known brands in our benchmark. To ensure fair comparison across varying difficulty match levels and baselines, we configure brand selection agent to output the predetermined brand for each prompt. Table~\ref{tab:known_brand_overall} presents results averaged across all test cases. BrandFusion achieves VBench quality scores comparable to baselines, indicating that brand integration does not compromise generation quality. However, BrandFusion significantly outperforms all baselines on semantic fidelity and naturalness scores, demonstrating that our framework successfully balances semantic preservation with natural integration. Notably, BrandFusion maintains high brand presence while achieving superior naturalness, indicating effective brand visibility without sacrificing scene coherence.
Single-LLM Rewriting lacks iterative refinement, resulting in lower naturalness. Template-based and Direct Append methods achieve high brand presence but often produce abrupt results.

\begin{table}[t]
    \centering
    \caption{Results on novel brands with model-level adaptation.}
    \label{tab:novel_brands}
    \resizebox{\columnwidth}{!}{%
    \begin{tabular}{@{}llcccc@{}}
        \toprule
        \textbf{Brand} & \textbf{T2V Model} & \textbf{CLIPScore} & \textbf{LLMScore} & \textbf{BPR} & \textbf{NS} \\ \midrule
        \multirow{3}{*}{ARUA} & Wan2.1-T2V-1.3B & 0.2934 & 0.9134 & 0.9312 & 3.94 \\
         & Wan2.2-T2V-5B & 0.3056 & 0.9487 & 0.9556 & 4.18 \\
         & CogVideoX1.5-5B & 0.2789 & 0.8923 & 0.8745 & 3.27 \\ \midrule
        \multirow{3}{*}{FreshWave} & Wan2.1-T2V-1.3B & 0.2898 & 0.9178 & 0.8267 & 3.98 \\
         & Wan2.2-T2V-5B & 0.3012 & 0.9523 & 0.9312 & 4.39 \\
         & CogVideoX1.5-5B & 0.2723 & 0.8867 & 0.7723 & 3.19 \\ \bottomrule
        \end{tabular}%
    }
\end{table}

\noindent\textbf{Results on Novel Brands.}
We evaluate the effectiveness of brand adaptation for two novel brands. As shown in Table~\ref{tab:novel_brands}, Wan2.2-T2V-5B consistently achieves the best performance with highest BPR and NS scores, attributed to its larger model capacity and stronger generation capabilities, while Wan2.1-T2V-1.3B maintains reasonable effectiveness despite its smaller size. Notably, all three models achieve high brand presence rates and naturalness scores, validating our adapter training's effectiveness. The consistent performance across diverse brand categories (footwear vs. beverage) confirms the generalizability of our synthetic data generation and LoRA fine-tuning approach, successfully enabling T2V models to generate novel brands with visual fidelity and natural scene integration even when entirely absent from pre-training data.

\begin{figure}[!tbp]
    \centering
    \includegraphics[width=1\columnwidth]{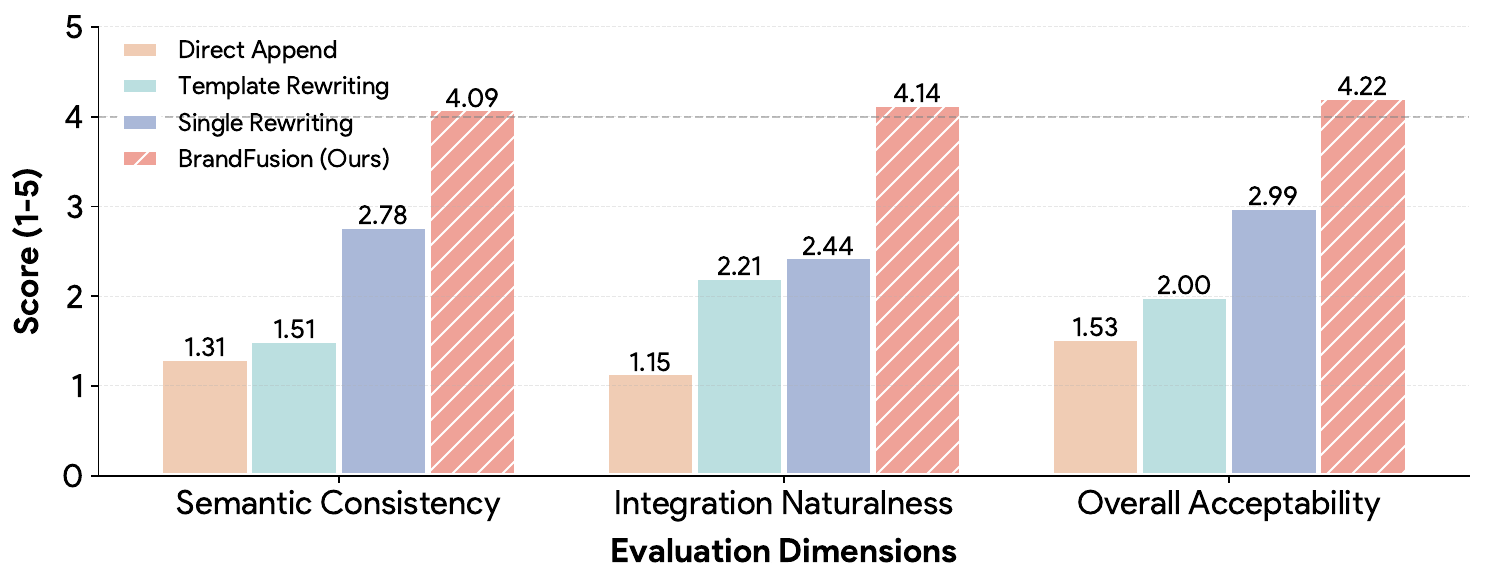}
    \caption{Human evaluation results on semantic fidelity, integration naturalness, and overall acceptability.}
    \label{fig:human_eval}
    \vspace{-6mm}
\end{figure}

\noindent\textbf{Human Evaluation.}
To validate our approach from the user perspective, we conduct a user study with 10 participants evaluating videos generated by the Veo3 model\footnote{This study was approved by our institution's  IRB.}. Participants assess videos across three dimensions using a 1-5 Likert scale: (1) \textit{Semantic Fidelity} --- preservation of original prompt intent; (2) \textit{Integration Naturalness} --- how naturally the brand is incorporated; and (3) \textit{Overall Acceptability} --- satisfaction with the generated video.
As shown in Figure~\ref{fig:human_eval} , BrandFusion consistently achieves the highest scores across all dimensions. This confirms that our framework successfully balances brand integration with user satisfaction, while baseline methods compromise user experience through semantic deviation or unnatural placements.

\section{Analysis}
\label{sec:analysis}

\subsection{Performance Across Different Match Level}

To assess robustness across varying scene-brand compatibility, we analyze performance stratified by match level. Table~\ref{tab:match_levels} reveals that BrandFusion maintains consistently strong performance across all compatibility levels, achieving high semantic fidelity and naturalness even in challenging Low Match scenarios. In contrast, baseline methods exhibit sharp degradation: Template-based Rewriting drops from NS 4.01 in high match to 1.38 in low match, while Direct Append shows significant decline in low match cases. BrandFusion gracefully handles difficult scenarios, maintaining strong semantic preservation while finding creative, context-aware integration strategies. This advantage validates the effectiveness of iterative multi-agent collaboration over one-pass optimization or fixed heuristics.

\subsection{Performance Across Prompt Scene Categories}

\begin{figure}[!t]
    \centering
    \includegraphics[width=1\columnwidth]{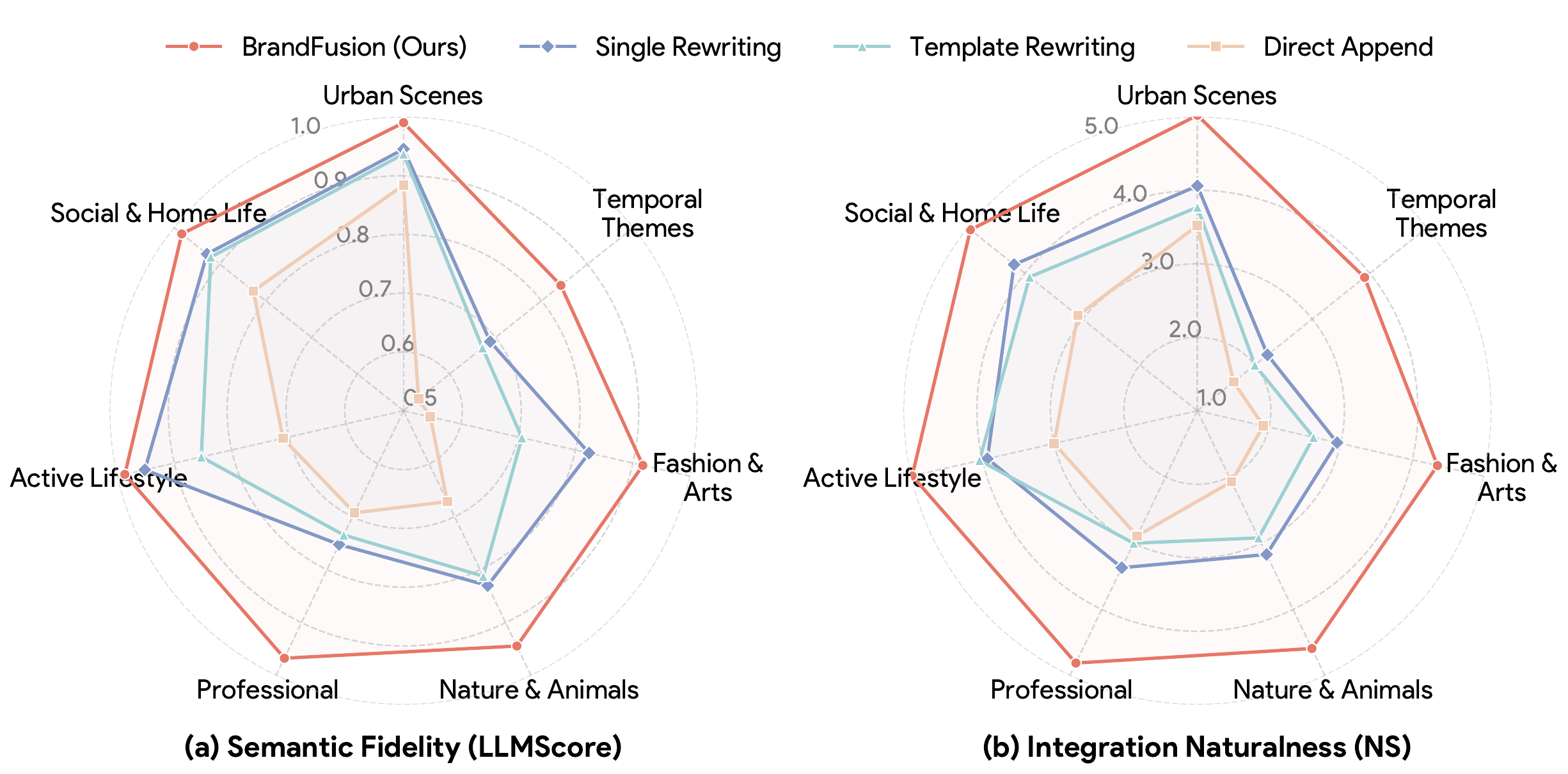}
    \caption{Comparison across different prompt scene categories.}
    \label{fig:scene_radar}
\end{figure}

To evaluate BrandFusion's adaptability across diverse real-world scenarios, we analyze performance on seven major scene categories aggregated from our 14 prompt scene types generated by Veo3. As illustrated in Figure~\ref{fig:scene_radar}, BrandFusion demonstrates consistent superiority over all baseline methods. Our method achieves particularly strong results in everyday contexts such as Urban Scenes, Social \& Home Life, where natural brand placement opportunities are abundant. While Temporal Themes, such as Sci-Fi and Historical scenes, present greater challenges due to the need to reconcile modern brands with temporally distinct contexts, BrandFusion still outperforms all baselines. These results validate that our framework enables effective cross-category generalization, maintaining high-quality brand integration regardless of scene complexity.

\begin{table}[!tbp]
    \centering
    \caption{Performance across different prompt-brand match levels.}
    \label{tab:match_levels}
    \resizebox{\columnwidth}{!}{%
    \begin{tabular}{@{}lcc|cc|cc@{}}
\toprule
 & \multicolumn{2}{c}{\textbf{High Match}} & \multicolumn{2}{c}{\textbf{Medium Match}} & \multicolumn{2}{c}{\textbf{Low Match}} \\ \cmidrule(l){2-7} 
\multirow{-2}{*}{\textbf{Method}} & LLMScore & \multicolumn{1}{c|}{NS} & LLMScore & \multicolumn{1}{c|}{NS} & LLMScore & NS \\ \midrule
Direct Append & 0.8234 & \multicolumn{1}{c|}{3.95} & 0.7876 & \multicolumn{1}{c|}{3.21} & 0.7353 & 1.33 \\
Template Rewriting & 0.9456 & \multicolumn{1}{c|}{4.01} & 0.9178 & \multicolumn{1}{c|}{3.97} & 0.9068 & 1.38 \\
Single Rewriting & 0.9512 & \multicolumn{1}{c|}{4.68} & 0.9489 & \multicolumn{1}{c|}{4.05} & 0.9235 & 2.97 \\
\rowcolor[HTML]{EFEFEF} 
BrandFusion (Ours) & 0.9734 & 4.90 & 0.9601 & 4.78 & 0.9333 & 4.42 \\ \bottomrule
        \end{tabular}%
    }
\end{table}

\subsection{Performance Across Brand Categories}

\begin{figure}[!t]
    \centering
    \includegraphics[width=1\columnwidth]{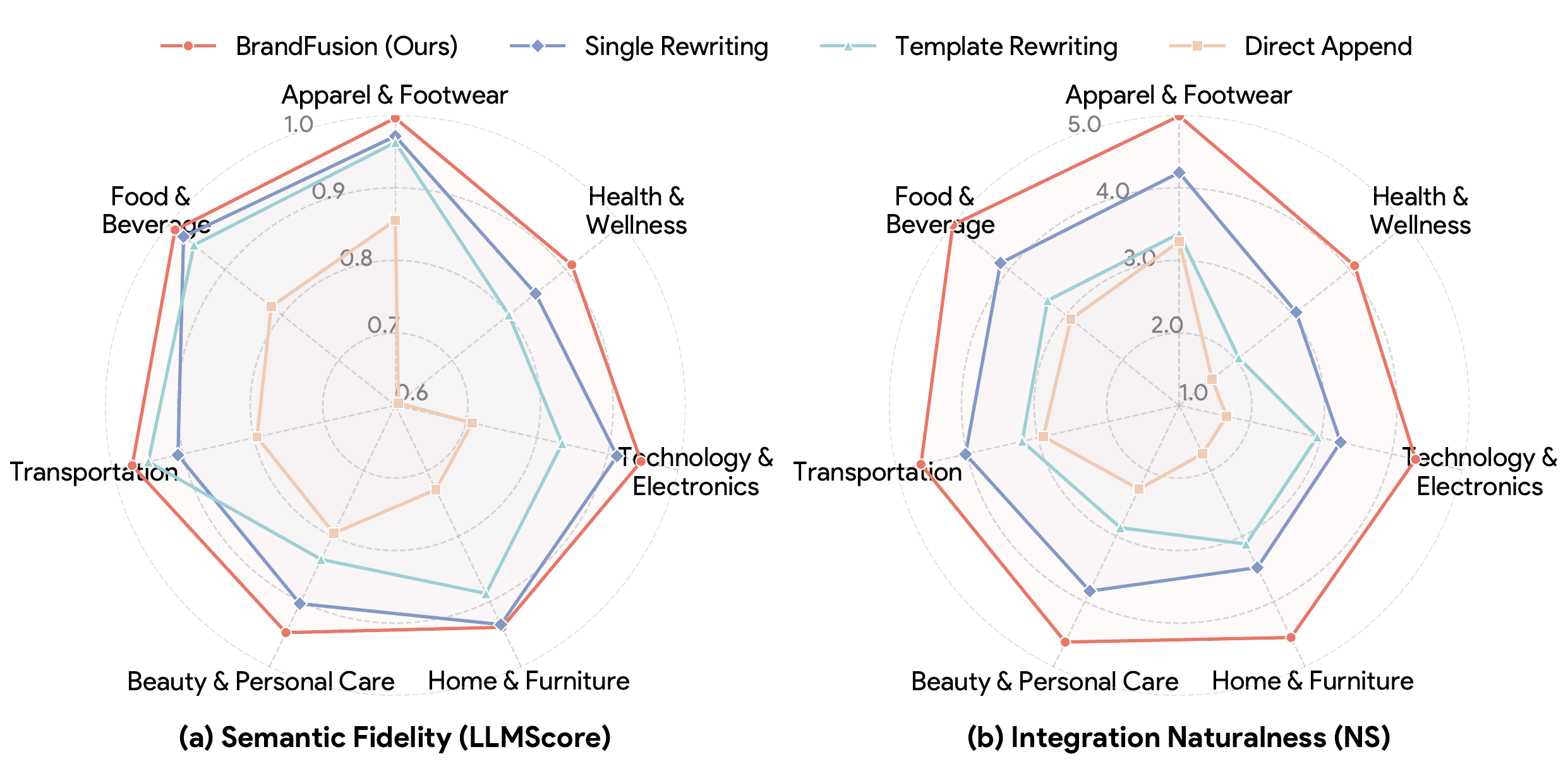}
    \caption{Comparison across seven brand categories.}
    \label{fig:brand_categories}
\end{figure}

To evaluate BrandFusion's generalizability across commercial domains, we analyze performance on seven brand categories spanning Food \& Beverage, Technology, Transportation, Apparel, Beauty, Home, and Health \& Wellness. As shown in Figure~\ref{fig:brand_categories}, BrandFusion consistently outperforms all baselines in both semantic fidelity and integration naturalness. Notably, everyday brands such as Apparel \& Footwear achieve the highest integration quality, attributed to their natural association with human subjects and flexible placement across diverse scenarios. Baseline methods show considerable performance variation, with Direct Append particularly struggling in specialized domains. While categories like Technology and Transportation present greater integration challenges due to more constrained contextual requirements, BrandFusion maintains relative higher performance, validating the effectiveness of our adaptive multi-agent framework across diverse commercial domains.
\begin{figure*}[!t]
    \centering
    \includegraphics[width=0.85\textwidth]{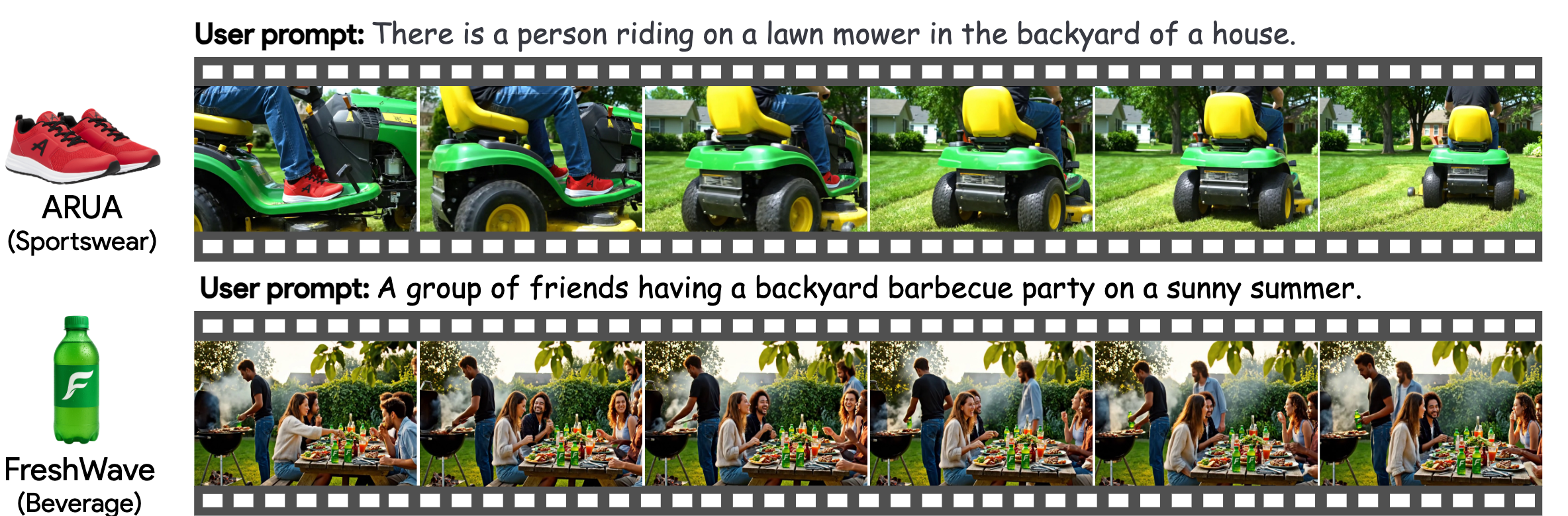}
    \caption{Case study examples of brand integration for novel brands using Wan2.2-T2V-5B. ARUA sportswear (top) and FreshWave beverage (bottom) are seamlessly integrated into diverse scenarios, demonstrating natural brand placement while preserving user intent.}
    \label{fig:case_study}
\end{figure*}

\subsection{Experience Learning Effectiveness}
\begin{figure}[t]
    \centering
    \includegraphics[width=1\columnwidth]{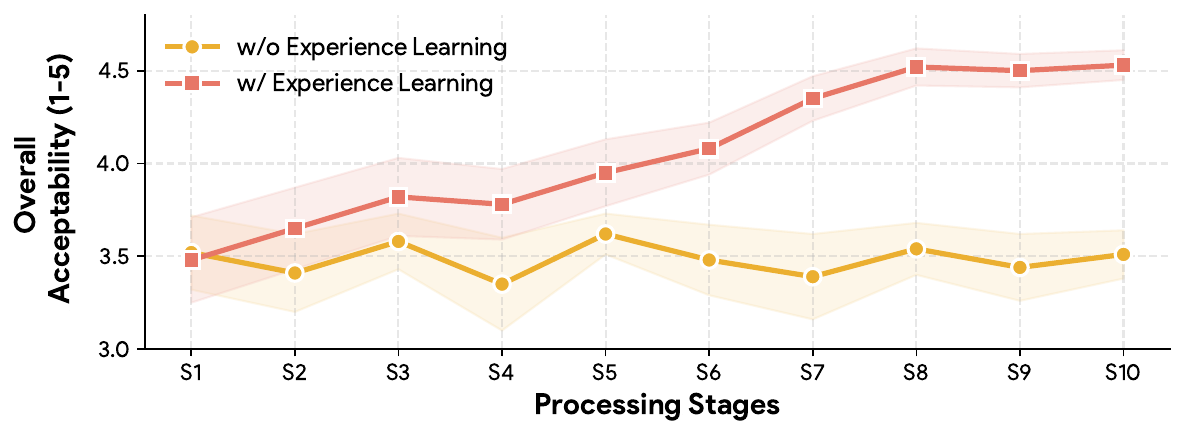}
    \caption{Experience learning effectiveness over sequential stages.}
    \label{fig:experience_learning}
\end{figure}

To validate the effectiveness of experience learning mechanism, we construct 100 diverse prompts for BMW brand integration, process them sequentially, and divide them into 10 groups to compute average overall acceptability within each group. As illustrated in Figure~\ref{fig:experience_learning}, BrandFusion with experience learning shows a clear upward trend across stages, while the baseline without experience learning remains relatively flat, demonstrating that the Experience Learning Agent successfully accumulates knowledge from previous integrations and continuously refines strategies.

\subsection{Efficiency Analysis}

While BrandFusion employs a multi-agent framework with iterative refinement, the online optimization process maintains reasonable efficiency. Our analysis reveals that the framework requires an average of 7.4 LLM calls per prompt, with an average latency of 16 seconds. This overhead is acceptable compared to actual video generation: Wan2.2-T2V-5B requires at least 120 seconds to generate a single video on an A100 GPU, meaning prompt optimization accounts for about 11\% of the pipeline. Combined with the substantial quality improvements demonstrated in our experiments, this makes BrandFusion viable for production deployment where video generation remains the dominant latency factor.

\subsection{Case Studies}

Figure~\ref{fig:case_study} illustrates representative examples of brand integration for novel brands using Wan2.2-T2V-5B. For the ARUA sportswear brand, the red shoes appear naturally on the rider's feet in a lawn mowing scene, maintaining clear visibility while preserving the activity context. For FreshWave beverage, the green bottle integrates seamlessly on the table during a backyard barbecue gathering, appearing as an organic element among other items. Both examples demonstrate BrandFusion's capability to handle novel brands absent from pre-training data, achieving recognizable brand presence while maintaining semantic fidelity and visual naturalness. More examples are shown in Appendix.

\section{Conclusion}
\label{sec:conclusion}
In this paper, we introduce seamless brand integration for text-to-video generation, a novel task that embeds brands into videos while preserving semantic fidelity to the user's original intent.
We propose BrandFusion, a multi-agent framework featuring offline brand knowledge construction and online collaborative prompt refinement through five specialized agents. Extensive experiments on 18 established and 2 novel brands across multiple state-of-the-art T2V models demonstrate that BrandFusion significantly outperforms baselines, with human evaluations confirming superior user satisfaction across diverse scenarios and brand categories. 
Beyond technical contributions, this work establishes a practical pathway for sustainable T2V monetization, enabling service providers to generate revenue, advertisers to achieve organic exposure, and users to receive high-quality content without disruptive interruptions. Future work includes exploring multi-brand integration and user-adaptive personalization strategies.

{
    \small
    \bibliographystyle{ieeenat_fullname}
    \bibliography{main}
}

\clearpage
\setcounter{page}{1}
\maketitlesupplementary

\startcontents[appendix]

{
\centering
\large\bfseries Appendix Contents\par
\vspace{1em}
}

\printcontents[appendix]{}{1}{\setcounter{tocdepth}{2}}

\appendix
\section{Dataset and Benchmark Details}
\label{app:dataset}

\subsection{Known Brand Benchmark}
\label{app:known_brands}
\begin{figure}[htbp]
    \centering
    \includegraphics[width=\columnwidth]{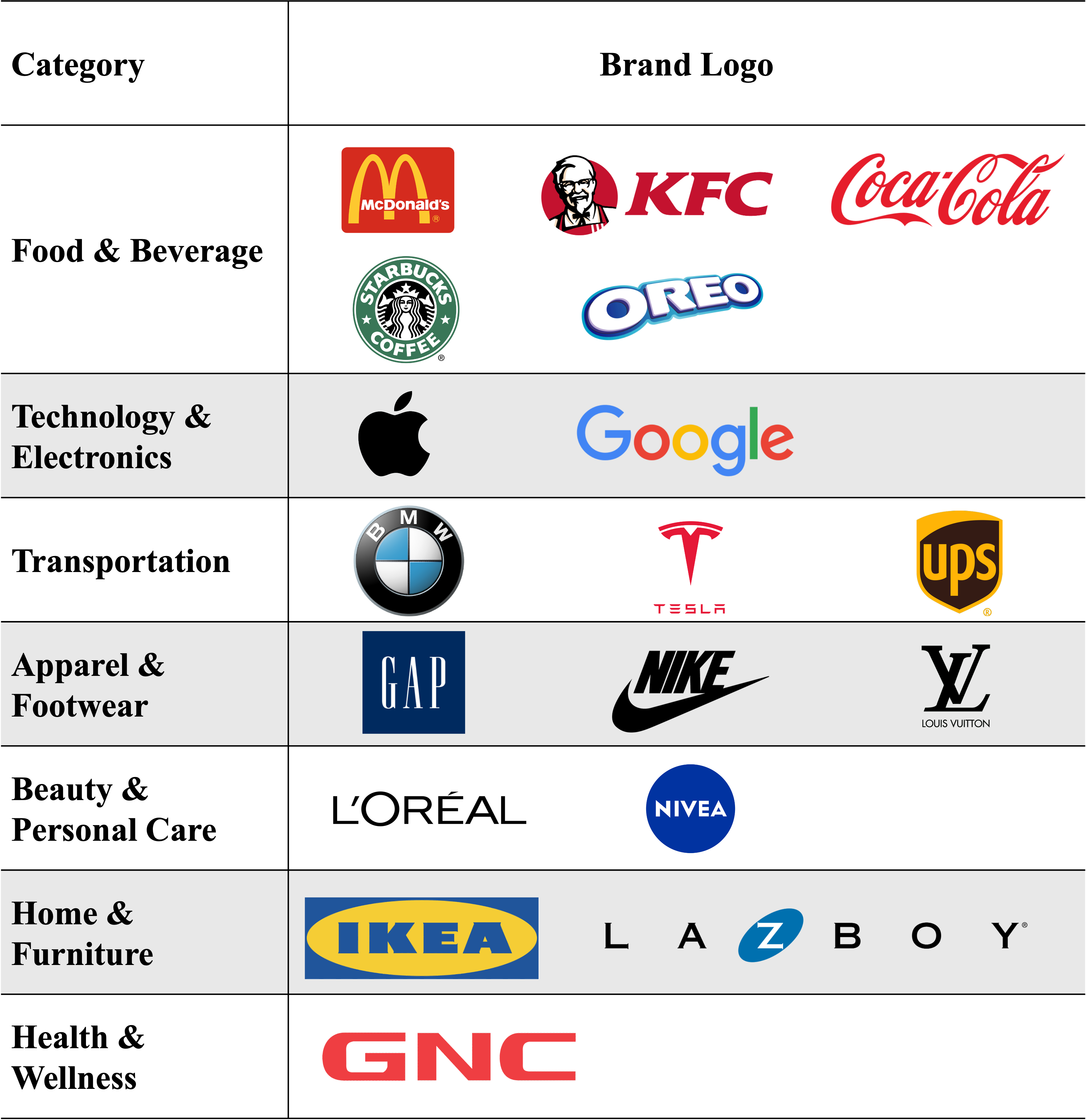}
    \caption{Complete list of 18 well-established brands across 7 industry categories used in our benchmark.}
    \label{fig:all_brands}
\end{figure}
To comprehensively evaluate BrandFusion's integration capabilities across diverse commercial domains, we curate a benchmark comprising 18 well-established brands spanning 7 major industry categories. These brands were selected to represent a wide spectrum of product types, market positions, and visual characteristics, ensuring our evaluation captures the framework's generalizability across different commercial contexts.
Figure~\ref{fig:all_brands} presents the complete list of brands organized by category:
\begin{itemize}
    \item \textbf{Food \& Beverage}: McDonald's, KFC, Coca-Cola, Starbucks, Oreo
    \item \textbf{Technology \& Electronics}: Apple, Google
    \item \textbf{Transportation}: BMW, Tesla, UPS
    \item \textbf{Apparel \& Footwear}: GAP, Nike, Louis Vuitton
    \item \textbf{Beauty \& Personal Care}: L'Oréal, NIVEA
    \item \textbf{Home \& Furniture}: IKEA, La-Z-Boy
    \item \textbf{Health \& Wellness}: GNC
\end{itemize}

\subsection{User Prompt Examples}
\label{app:prompts}

For each of the brands, we  construct 15 diverse user prompts that span varying levels of prompt-brand compatibility and cover different scene types. These prompts are categorized into three match levels:
\begin{itemize}
    \item \textbf{High Match}: Scenarios where the brand naturally fits and would typically appear. These represent ideal integration contexts.
    \item \textbf{Medium Match}: Scenarios where the brand can reasonably appear but requires more creative integration. These test the framework's adaptability.
    \item \textbf{Low Match}: Challenging scenarios with low semantic relevance, requiring sophisticated context-aware strategies. These evaluate the framework's robustness in difficult cases.
\end{itemize}
Table~\ref{tab:prompt_examples} presents representative examples for Coca-Cola across all three match levels and diverse scene categories, illustrating the range of scenarios our benchmark encompasses.

\begin{table*}[htbp]
    \centering
    \caption{User prompt examples for Coca-Cola brand across different match levels and scene categories. Each prompt represents a distinct integration challenge, from natural placement scenarios (high match) to creative adaptation contexts (low match).}
    \label{tab:prompt_examples}
    \resizebox{0.85\textwidth}{!}{%
    \begin{tabular}{@{}lp{9cm}l@{}}
        \toprule
        \textbf{Scene Category} & \textbf{User Prompt} & \textbf{Match Level} \\ 
        \midrule
        \multicolumn{3}{l}{\textit{High Match Scenarios}} \\
        \midrule
        Social \& Leisure & A group of friends sharing drinks at a backyard barbecue. & High \\
        Indoor \& Home Life & A family enjoying snacks and beverages during a movie night at home. & High \\
        Nature \& Outdoors & A woman relaxing on a sunny beach with a cold drink in hand. & High \\
        Work \& Business & A teenager buying a cold soda from a convenience store vending machine. & High \\
        Social \& Leisure & A couple sharing a meal and drinks at a cozy diner. & High \\
        \midrule
        \multicolumn{3}{l}{\textit{Medium Match Scenarios}} \\
        \midrule
        Education \& Academia & A college student studying late in the library with snacks nearby. & Medium \\
        Urban \& Streetscape & A busy city street with people walking past colorful billboards. & Medium \\
        Work \& Business & Office workers chatting during a lunch break in the break room. & Medium \\
        Travel \& Transportation & A taxi driver taking a break at a busy intersection. & Medium \\
        Nature \& Outdoors & Children playing on swings at a neighborhood playground. & Medium \\
        \midrule
        \multicolumn{3}{l}{\textit{Low Match Scenarios}} \\
        \midrule
        Sports \& Fitness & A yoga instructor stretching in a serene park at sunrise. & Low \\
        Pets \& Animals & A dog waiting patiently beside a picnic blanket in the countryside. & Low \\
        Fashion \& Beauty & A fashion designer sketching new ideas in a stylish studio. & Low \\
        Historical \& Period & A vintage steam train crossing through a mountain valley. & Low \\
        Urban \& Streetscape & A bustling street in a cyberpunk city at night, with neon signs reflecting on the wet pavement. & Low \\
        \bottomrule
    \end{tabular}%
    }
\end{table*}

\section{Implementation Details}
\label{app:implementation}

\subsection{Brand Knowledge Base Structure}
\label{app:knowledge_base}

The Brand Knowledge Base serves as the centralized long-term memory system that stores comprehensive information about all registered brands. It is implemented as a structured database with the following schema for each brand entry:

\begin{itemize}
    \item \textbf{Brand Profile}: Core metadata including brand name $\mathcal{N}$, category $\mathcal{C}$ (\textit{e.g.}, Food \& Beverage, Technology), brand description $\mathcal{D}$, and reference images $\mathcal{R}$.
    
    \item \textbf{Knowledge Type}: A binary flag indicating whether the brand has sufficient prior knowledge in the T2V model (\texttt{has\_prior\_knowledge}: True/False). This determines whether an adapter is required during generation.
    
    \item \textbf{Brand Adapter}: For brands without prior knowledge, we store the path to the trained LoRA adapter weights $\mathcal{A}_{\mathcal{B}}$ and associated metadata (training configuration, trigger token).
    
    \item \textbf{Visual Patterns}: A collection of reference visual elements including Logo images and product images.
    
     \item \textbf{Experience Pool}: A continuously updated collection of abstracted integration experiences learned from past successful and failed cases. Each experience is stored as a natural language pattern that captures generalizable insights, such as:\textit{``Outdoor sports scenes are highly suitable for athletic footwear brand integration, especially when athletes or active individuals are present.''}; \textit{``Urban street scenes with billboards provide natural placement opportunities for beverage brands without disrupting the scene.''}

    \item \textbf{Prohibited Scenes}: A list of scene types or contexts where the brand should not appear (\textit{e.g.}, violent scenes, inappropriate contexts) based on brand guidelines and ethical considerations.
\end{itemize}

\subsection{Algorithm: Multi-Agent Brand Integration}
\label{app:algorithm}
This section presents the complete algorithmic workflow of BrandFusion's online phase, detailing how five specialized agents collaborate to achieve seamless brand integration. Algorithm~\ref{alg:main} describes the overall multi-agent integration process, while Algorithm~\ref{alg:refinement} details the iterative refinement loop.

\begin{algorithm*}[htbp]
\caption{Online Multi-Agent Brand Integration}
\label{alg:main}
\begin{algorithmic}[1]
\REQUIRE User prompt $\mathcal{P}_u$, Brand Knowledge Base $\mathcal{KB}$, T2V model $\mathcal{G}_\theta$
\ENSURE Generated video $\mathcal{V}$ with brand integration
\STATE Initialize Working Context $\mathcal{WC} \leftarrow \{\mathcal{P}_u, \text{iteration} = 0, \text{history} = []\}$
\STATE \textcolor{blue}{// Phase 1: Brand Selection}
\STATE $\mathcal{B}^* \leftarrow$ \textsc{BrandSelectionAgent}($\mathcal{P}_u, \mathcal{KB}$)
\STATE Retrieve brand profile: $\{\mathcal{N}, \mathcal{C}, \mathcal{R}, \mathcal{D}, \text{has\_prior}, \mathcal{A}_{\mathcal{B}}, \text{experiences}\}$ from $\mathcal{KB}$
\STATE Update $\mathcal{WC}$.selected\_brand $\leftarrow \mathcal{B}^*$
\STATE \textcolor{blue}{// Phase 2: Iterative Refinement}
\STATE $\mathcal{P}_{\text{opt}}, \text{converged} \leftarrow$ \textsc{IterativeRefinement}($\mathcal{P}_u, \mathcal{B}^*, \mathcal{KB}, \mathcal{WC}$)
\IF{not converged}
    \RETURN Error: "Failed to generate acceptable prompt"
\ENDIF
\STATE \textcolor{blue}{// Phase 3: Video Generation}
\IF{$\mathcal{B}^*$.has\_prior = False}
    \STATE Load brand adapter $\mathcal{A}_{\mathcal{B}}$ into $\mathcal{G}_\theta$
\ENDIF
\STATE $\mathcal{V} \leftarrow \mathcal{G}_\theta(\mathcal{P}_{\text{opt}})$ \textcolor{gray}{// Generate video}
\STATE \textcolor{blue}{// Phase 4: Experience Learning}
\STATE Collect user feedback $f$ (thumbs up/down, ratings)
\STATE \textsc{ExperienceLearningAgent}($\mathcal{WC}, f, \mathcal{KB}$)
\RETURN $\mathcal{V}$
\end{algorithmic}
\end{algorithm*}

\begin{algorithm*}[htbp]
\caption{Iterative Refinement Loop}
\label{alg:refinement}
\begin{algorithmic}[1]
\REQUIRE User prompt $\mathcal{P}_u$, Selected brand $\mathcal{B}^*$, Knowledge Base $\mathcal{KB}$, Working Context $\mathcal{WC}$, $\text{max\_iterations}$
\ENSURE Optimized prompt $\mathcal{P}_{\text{opt}}$, convergence flag
\STATE $\mathcal{P}' \leftarrow \mathcal{P}_u$ \textcolor{gray}{// Initialize with user prompt}
\FOR{$i = 1$ to max\_iterations}
    \STATE $\mathcal{WC}.\text{iteration} \leftarrow i$
    \STATE \textcolor{blue}{// Step 1: Strategy Generation}
    \STATE Query relevant experiences from $\mathcal{KB}$.experience\_pool based on scene similarity
    \STATE $\mathcal{S} \leftarrow$ \textsc{StrategyGenerationAgent}($\mathcal{P}_u, \mathcal{B}^*, \text{experiences}, \mathcal{WC}$)
    \STATE \textcolor{blue}{// Step 2: Prompt Rewriting}
    \STATE $\mathcal{P}' \leftarrow$ \textsc{PromptRewritingAgent}($\mathcal{P}_u, \mathcal{B}^*, \mathcal{S}, \mathcal{WC}$)
    \STATE \textcolor{blue}{// Step 3: Critic Evaluation}
    \STATE $\text{decision}, \text{feedback} \leftarrow$ \textsc{CriticAgent}($\mathcal{P}', \mathcal{P}_u, \mathcal{B}^*, s_{\text{sem}}, s_{\text{brand}}, s_{\text{nat}}, s_{\text{qual}}$)
    \STATE \textcolor{blue}{// Store iteration history}
    \STATE Append $\{\mathcal{S}, \mathcal{P}', \text{scores}, \text{decision}, \text{feedback}\}$ to $\mathcal{WC}.\text{history}$
    \STATE \textcolor{blue}{// Step 4: Decision Making}
    \IF{decision = ``accept''}
        \STATE $\mathcal{WC}.\text{final\_prompt} \leftarrow \mathcal{P}'$
        \RETURN $\mathcal{P}'$, True \textcolor{gray}{// Success}
    \ELSIF{decision = ``revise''}
        \STATE Continue to next iteration with feedback
    \ELSIF{decision = ``replan''}
        \STATE Discard current strategy $\mathcal{S}$
        \STATE Mark current approach as failed in $\mathcal{WC}$
        \STATE Continue to next iteration to generate new strategy
    \ENDIF
\ENDFOR
\RETURN $\mathcal{P}'$, False \textcolor{gray}{// Failed to converge within max iterations}
\end{algorithmic}
\end{algorithm*}

\section{Agent Prompt Templates}
\label{app:agent_prompts}

This section provides the complete prompt templates used for each of the five specialized agents in BrandFusion's online multi-agent framework. Each agent is powered by a large language model (GPT-5) and receives carefully designed prompts that define its role, task, input format, and expected output format. The prompts incorporate dynamic variables (shown in \textcolor{blue}{blue brackets}) that are populated at runtime with context-specific information from the Brand Knowledge Base and Working Context. 

\subsection{Brand Selection Agent}
\label{app:prompt_selection}

The Brand Selection Agent serves as the entry point of the multi-agent workflow. Its primary responsibility is to analyze the user's prompt and select the most compatible brand from the Brand Knowledge Base.

\begin{graybox}{Brand Selection Agent Prompt Template}
\small
You are a Brand Selection Agent specializing in identifying the most suitable brand for seamless integration into user-requested video scenarios. 

Given a user prompt describing a desired video scene, analyze the scene characteristics and select the most compatible brand from the available Brand Knowledge Base that can be naturally integrated while preserving the user's creative intent.

\textbf{Input Information:}

- \textbf{User Prompt:} \textcolor{blue}{\{user\_prompt\}}

- \textbf{Available Brands:} \textcolor{blue}{\{brand\_knowledge\_base\}}

Each brand contains:
- Brand Name

- Category

- Prohibited Scenarios

\textbf{Selection Criteria:}

1. \textbf{Semantic Compatibility:} Assess how naturally the brand fits within the described scene context. Consider the scene type, setting, characters, and activities.

2. \textbf{Category Relevance:} Evaluate whether the brand's product category aligns with typical usage scenarios in the prompt.

3. \textbf{Visual Feasibility:} Determine if there are natural placement opportunities for the brand (e.g., products, logos, branded objects) within the scene.

4. \textbf{Avoided Scenarios:} Ensure the prompt does not fall into the brand's prohibited scenarios that would result in inappropriate or unsuccessful integration.

\textbf{Output Format:}

Provide your selection in a structured JSON format including: selected brand name, compatibility score (0.0-1.0), reasoning for selection.

\textbf{Important Guidelines:}

- Prioritize brands that can be integrated naturally without forcing placement.

- Consider multiple potential integration points within the scene.

- Be mindful of maintaining the user's original creative intent.

- If multiple brands have similar compatibility, select the one with the strongest visual presence potential.

- Provide clear reasoning to support your selection for transparency.

Now, analyze the user prompt and select the most appropriate brand.
\end{graybox}

\subsection{Strategy Generation Agent}
\label{app:prompt_strategy}

The Strategy Generation Agent is responsible for designing context-aware integration strategies that balance semantic preservation with brand visibility.

\begin{graybox}{Strategy Generation Agent Prompt Template}
\small
You are a strategic planning specialist for seamless brand integration in video generation. 
Your mission is to create innovative and effective strategies for naturally integrating brand elements into users' video generation prompts while preserving their original creative intent and ensuring the brand appears organically within the scene context.

\textbf{Input Information:}

- \textbf{User's Original Prompt:} \textcolor{blue}{\{user\_prompt\}}

- \textbf{Selected Brand:} \textcolor{blue}{\{selected\_brand\}}

- \textbf{Brand Category:} \textcolor{blue}{\{brand\_category\}}

- \textbf{Previous Strategies Tried:} \textcolor{blue}{\{previous\_strategies\}}

- \textbf{Experience Pool - Successful Cases:} \textcolor{blue}{\{successful\_experiences\}}

Your task is generating a concise integration strategy that naturally fits the scene context. The strategy should specify HOW the brand will be integrated (placement approach, form of appearance) rather than providing detailed scene descriptions. Focus on strategic thinking that ensures natural brand presence without disrupting the user's creative vision.

You may refer to but do not be limited to the following possible strategies.

- \textbf{Main Object Integration:} Brand product serves as the primary subject or functional element

- \textbf{Background Elements:} Brand appears naturally in the environment (on shelves, tables, walls, posters, billboards)

- \textbf{Character Interaction:} People in the scene use, wear, or hold brand items

- \textbf{Environmental Details:} Brand elements become part of the setting without being the main focus

- \textbf{Lifestyle Representation:} Brand reflects the lifestyle or values of scene participants

- \textbf{Contextual Placement:} Unexpected but natural appearances that enhance scene authenticity

- \textbf{Ambient Integration:} Brand subtly present through environmental cues or secondary objects

\textbf{Strategy Requirements:}

- Keep strategy concise (1-2 sentences maximum)

- Ensure strategy differs from previous attempts that were unsuccessful

- Learn from successful integration patterns in similar scene contexts

- Avoid approaches that have failed in the experience pool

- Consider the specific context and constraints of the user's original prompt

- Prioritize naturalness over aggressive brand visibility

- Ensure the integration aligns with the scene's mood, setting, and narrative flow

\textbf{Output Format:}

Provide your response in JSON format with a single field 'strategy' containing one concise integration strategy.

Now, analyze the scene context and generate an effective integration strategy.
\end{graybox}

\subsection{Prompt Rewriting Agent}
\label{app:prompt_rewriting}

The Prompt Rewriting Agent transforms the user's original prompt into an optimized video generation prompt that seamlessly integrates the selected brand while preserving semantic fidelity. 

\begin{graybox}{Prompt Rewriting Agent Prompt Template}
\small
You are a world-class video prompt specialist for seamless brand integration. Your mission is to transform users' original prompts into comprehensive video generation prompts that artfully integrate specific brands while preserving the original creative intent and ensuring natural scene coherence.

\textbf{Input Information:}

- \textbf{User's Original Prompt:} \textcolor{blue}{\{user\_prompt\}}

- \textbf{Selected Brand:} \textcolor{blue}{\{selected\_brand\}}

- \textbf{Brand Category:} \textcolor{blue}{\{brand\_category\}}

- \textbf{Integration Strategy:} \textcolor{blue}{\{integration\_strategy\}}

- \textbf{Target Video Duration:} \textcolor{blue}{\{video\_duration\}} seconds

- \textbf{Previous Revision Feedback:} \textcolor{blue}{\{revision\_history\}}

\textbf{Core Integration Principles:}

1. \textbf{Semantic Preservation:} Maintain the user's original creative intent, including key subjects, actions, mood, and stylistic preferences. Do not alter or destroy the fundamental meaning of the user's prompt.

2. \textbf{Natural Integration:} The brand should appear as an organic part of the scene, not as the primary focus or dominant element. Brand visibility should be sufficient for recognition but balanced with scene authenticity.

3. \textbf{Logical Consistency:} Ensure the scene is self-consistent and plausible, following real-world logic (physics, action continuity, object permanence, lighting consistency) and aligning with the provided integration strategy.

4. \textbf{Style Consistency:} Follow professional video generation prompt conventions, maintaining conciseness and clarity while avoiding unnecessary complexity.

\textbf{Brand Integration Guidelines:}

- The brand must be naturally integrated according to the given strategy

- Brand elements should be recognizable and noticeable but not dominate the scene

- The brand should NOT be the main subject or primary focus of the video

- Integration should feel like a natural part of the environment or character interactions

- Viewers should be able to identify the brand while scene authenticity is maintained

\textbf{Prompt Design Structure:}

A high-quality video generation prompt should be concise, descriptive, and structured as a single coherent paragraph. Consider including:

- \textbf{Subject:} Main object, person, animal, or scenery

- \textbf{Context:} Background or environment setting

- \textbf{Action:} What subjects are doing (walking, running, interacting, etc.)

- \textbf{Style:} Film style keywords (cinematic, documentary, animation style, etc.)

- \textbf{Camera Motion:} [Optional] Positioning and movement (aerial view, tracking shot, dolly shot, pan, zoom)

- \textbf{Composition:} [Optional] Shot framing (wide shot, close-up, medium shot, over-the-shoulder)

- \textbf{Visual Effects:} [Optional] Focus and lens (shallow focus, macro lens, wide-angle, bokeh)

- \textbf{Ambiance:} [Optional] Lighting and color tone (warm tones, golden hour, blue hour, dramatic lighting)

\textbf{Important Constraints:}

- Keep the prompt concise and avoid excessive complexity

- The number of shots should be realistic for the target video duration

- Do NOT explicitly specify duration in the prompt text

- Do NOT add unnecessary details unrelated to the user's original intent

- Ensure all scene elements follow real-world plausibility

- The integrated prompt must align with the provided integration strategy

Based on the integration strategy and all input information, your task is to write a complete video generation prompt that naturally incorporates the brand while faithfully preserving the user's original creative vision.

Now, generate the optimized video generation prompt.
\end{graybox}

\subsection{Critic Agent}
\label{app:prompt_critic}

The Critic Agent performs multi-dimensional evaluation of the rewritten prompt, assessing whether it successfully balances semantic fidelity, brand visibility, and integration naturalness.

\begin{graybox}{Critic Agent Prompt Template}
\small
You are an expert evaluator specializing in assessing video generation prompts for seamless brand integration. Your mission is to rigorously evaluate whether rewritten prompts meet high standards for both preserving user creative intent and achieving natural brand integration.

\textbf{Input Information:}

- \textbf{User's Original Prompt:} \textcolor{blue}{\{user\_prompt\}}

- \textbf{Selected Brand:} \textcolor{blue}{\{selected\_brand\}}

- \textbf{Brand Category:} \textcolor{blue}{\{brand\_category\}}

- \textbf{Integration Strategy:} \textcolor{blue}{\{integration\_strategy\}}

- \textbf{Target Video Duration:} \textcolor{blue}{\{video\_duration\}} seconds

- \textbf{Rewritten Prompt to Evaluate:} \textcolor{blue}{\{rewritten\_prompt\}}

- \textbf{Revision History:} \textcolor{blue}{\{revision\_history\}}

\textbf{Evaluation Dimensions:}

Assess the rewritten prompt across the following critical dimensions:

1. \textbf{Semantic Fidelity:} Does the rewritten prompt faithfully preserve the user's original creative intent? Evaluate whether the core idea, mood, key subjects, actions, and stylistic preferences remain intact. The prompt should NOT destroy or fundamentally alter the user's original meaning.

2. \textbf{Brand Clarity and Recognizability:} Is the brand element clearly identifiable and recognizable? The brand should be visible enough for viewers to identify it, with sufficient detail to ensure brand presence is not ambiguous or too subtle.

3. \textbf{Integration Naturalness:} Does the brand appear organically within the scene context? The brand should feel like a natural part of the environment or character interactions, not forced or artificially inserted. It should enhance rather than disrupt scene authenticity.

4. \textbf{Strategy Alignment:} Does the rewritten prompt properly execute the specified integration strategy? The prompt should follow the strategic guidance without deviating into unplanned approaches.

5. \textbf{Generation Effectiveness:} Is the described scene realistically achievable within the target video duration and technically feasible for T2V models? The scene complexity and shot count should be appropriate for the specified duration.

\textbf{Decision Guidelines:}

Based on your evaluation, make one of three decisions:

- \textbf{Accept:} The prompt successfully meets all quality standards. Semantic fidelity is preserved, brand integration is natural and recognizable, and the prompt is ready for video generation.

- \textbf{Revise:} The prompt has issues that can be fixed through refinement without changing the strategy. Provide specific, actionable feedback focusing on what needs improvement (e.g., semantic elements to restore, brand visibility adjustments, naturalness improvements).

- \textbf{Replan:} The current strategy is fundamentally flawed and cannot produce satisfactory results through revision alone. This should be chosen when the strategy itself causes inherent conflicts between semantic preservation and brand integration, or when multiple revision attempts have failed.

\textbf{Feedback Guidelines:}

- If accepting, confirm that quality standards are met
- If requesting revision, provide concise, actionable suggestions focusing on specific improvements needed
- If requesting replanning, explain why the current strategy is fundamentally problematic
- Focus on what needs to be improved, not how to improve it (that's the Prompt Rewriting Agent's job)
- Be constructive and specific rather than vague

\textbf{Output Format:}

Provide your response in JSON format with two fields: 'decision' (accept/revise/replan) and 'feedback' (specific evaluation feedback or improvement suggestions).

Now, conduct your evaluation and provide your decision.
\end{graybox}

\subsection{Experience Learning Agent}
\label{app:prompt_experience}

The Experience Learning Agent completes the closed-loop learning mechanism by collecting user feedback and abstracting integration outcomes into reusable experiences. 

\begin{graybox}{Experience Learning Agent Prompt Template}
\small
You are an experience learning specialist responsible for analyzing brand integration outcomes and extracting valuable insights for future integrations. Your mission is to learn from both successful and unsuccessful integration attempts, abstracting transferable patterns that can guide future brand placement strategies.

\textbf{Input Information:}

- \textbf{User's Original Prompt:} \textcolor{blue}{\{user\_prompt\}}

- \textbf{Selected Brand:} \textcolor{blue}{\{selected\_brand\}}

- \textbf{Brand Category:} \textcolor{blue}{\{brand\_category\}}

- \textbf{Integration Strategy Used:} \textcolor{blue}{\{integration\_strategy\}}

- \textbf{Final Rewritten Prompt:} \textcolor{blue}{\{final\_prompt\}}

- \textbf{User Feedback:} \textcolor{blue}{\{user\_feedback\}}

- \textbf{Outcome:} \textcolor{blue}{\{outcome\}} (success/failure)

\textbf{Your Task:}

Analyze the integration outcome and extract actionable insights that can inform future brand integration decisions. Identify patterns, successful approaches, or pitfalls to avoid, creating experiences that will guide future integrations in similar scenarios.

\textbf{Analysis Focus:}

For successful integrations:
- What made the integration natural and effective?
- Which strategy aspects worked particularly well?
- What scene characteristics facilitated successful placement?

For unsuccessful integrations:
- What caused the integration to feel forced or unnatural?
- Why did the strategy fail to achieve the desired balance?
- What should be avoided in similar future scenarios?

\textbf{Experience Requirements:}

- \textbf{Generalizability:} Abstract insights to apply beyond this specific case
- \textbf{Actionability:} Provide concrete guidance that other agents can use
- \textbf{Clarity:} Express insights concisely and clearly

\textbf{Output Format:}

Provide your analysis in JSON format with the following fields:
- 'experience type': success or failure
- 'key insight': The main lesson learned (1-2 sentences)

Now, analyze the integration outcome and extract valuable experience for the knowledge base.
\end{graybox}
\section{Evaluation Methodology Details}
\label{app:evaluation}

This section provides detailed descriptions of the evaluation methodologies used to assess BrandFusion's performance across multiple dimensions. We employ both automated metrics and human evaluation protocols to comprehensively measure video quality, semantic fidelity, and brand integration quality. The following subsections describe the implementation details of each evaluation approach, including the prompts used for LLM-based assessments.

\subsection{VQAScore Implementation}
\label{app:vqascore}

VQAScore measures semantic fidelity by evaluating whether the generated video preserves the key information from the user's original prompt. The methodology operates in two stages: (1) question generation, where we use an LLM to generate a set of questions based on the original prompt that cover all important aspects, and (2) answer verification, where we use a multimodal VQA model to answer these questions by watching the generated video and compare the answers against ground truth.

\paragraph{Question Generation Process:}
Given the user's original prompt, we employ an LLM to generate exactly 8 questions that comprehensively cover all aspects of the prompt. These questions focus on key entities, actions, attributes, relationships, and settings described in the original prompt. Each question is paired with its correct answer derived from the prompt, serving as ground truth for evaluation.

\paragraph{Answer Verification Process:}
For each generated video, we input the video frames along with the generated questions into a multimodal VQA model. The model answers each question based on what it observes in the video. We then compare the model's answers with the ground truth answers using semantic similarity metrics. The final VQAScore is computed as the average accuracy across all 8 questions, with scores ranging from 0 to 1, where higher scores indicate better semantic preservation.

The question generation prompt template is shown below:

\begin{graybox}{VQAScore Question Generation Prompt}
\small
You are an expert in video content evaluation. Given a text description, your task is to generate exactly 8 questions with their correct answers. These questions should be used to evaluate whether a video is semantically consistent with the text description.

\textbf{Requirements:}

- Generate exactly 8 questions

- Each question can focus on specific key points or cover multiple aspects

- Together, all 8 questions must cover ALL aspects of the text description to avoid missing any key information

- Questions should be answerable by watching the video

- Questions should be specific and unambiguous

- Provide the correct answer for each question based on the text description

\textbf{Text Description:}

\textcolor{blue}{\{user\_prompt\}}

\textbf{Question Guidelines:}

- Focus on key entities, actions, attributes, relationships, and settings

- Use clear, direct language

- Questions can be:

  * General questions like "Does the video show '\textcolor{blue}{\{text\}}'?" or "Is the scene consistent with '\textcolor{blue}{\{text\}}'?"
  
  * Yes/no questions about the presence of specific objects, actions, or attributes
  
  * Specific "what", "where", "how", "who" questions about particular elements
  
- Include at least one general question that evaluates overall consistency

- Include specific questions for each important detail in the text

- Ensure answers are concrete and verifiable from the text

- Do not ask questions that are unrelated to the text description

\textbf{Output Format:}

Provide your response in JSON format with a "questions" array, where each element contains "question" and "answer" fields for all 8 questions.

Now, generate the 8 evaluation questions.
\end{graybox}

\subsection{Naturalness Score Implementation}
\label{app:naturalness}

Naturalness Score (NS) assesses the quality of brand integration from three complementary perspectives: contextual fit, visual blend, and non-intrusiveness. This metric evaluates whether the brand appears organically within the scene rather than as a forced or artificial insertion. Unlike binary presence detection, NS provides fine-grained assessment on a 1-5 scale across three dimensions, with the final score computed as the average of the three sub-scores. 
The evaluation prompt template is shown below:

\begin{graybox}{Naturalness Score Evaluation Prompt}
\small
You are an expert evaluator specializing in visual media and brand integration. Your task is to analyze an AI-generated video and evaluate how naturally a target brand was integrated into the scene based on the original user prompt.

\textbf{Input Information:}

- \textbf{Original User Prompt:} \textcolor{blue}{\{user\_prompt\}}
- \textbf{Target Brand:} \textcolor{blue}{\{brand\_name\}}
- \textbf{Video:} \textcolor{blue}{\{video\_file\}}

\textbf{Evaluation Task:}

Evaluate the Natural Placement Score (NPS) for the brand in the video based on three key criteria. Watch the video carefully and assess each dimension independently.

\textbf{Evaluation Criteria:}

1. \textbf{Contextual Fit (Score 1-5):}
   - Is the brand logical and appropriate for the scene, characters, and activities described in the original prompt?
   - Example: A sports drink in a gym is contextually appropriate; a car tire on a dinner table is inappropriate.

2. \textbf{Visual Blend (Score 1-5):}
   - Does the brand element look like it visually belongs in the scene?
   - Evaluate: Lighting consistency, shadow accuracy, perspective and scale correctness, occlusion relationships, absence of artifacts (floating, pasted-on appearance, clipping through objects)

3. \textbf{Non-Intrusiveness (Score 1-5):}
   - Does the brand placement obstruct or distract from the main subject or action in the original prompt?
   - Is the brand naturally placed in the background or as a secondary object, or is it excessively prominent, making it the main focus?

\textbf{Scoring Rubric (1-5 Scale):}

- \textbf{5 (Excellent):} Seamless integration. Contextually perfect, visually indistinguishable from the scene, and does not intrude on the main subject.
- \textbf{4 (Good):} Well-integrated. Contextually appropriate and visually well-blended with only very minor flaws. Not intrusive.
- \textbf{3 (Average):} Acceptable. Contextually logical but has minor visual blend issues (e.g., slight lighting inconsistency) OR is slightly distracting.
- \textbf{2 (Bad):} Highly flawed. Either contextually inappropriate OR has major visual blend issues (looks pasted-on, obvious artifacts).
- \textbf{1 (Poor):} Unnatural. Contextually wrong, looks fake or pasted-on, AND is highly intrusive.

\textbf{Output Format:}

Provide your analysis in JSON format with four components: 'contextual\_fit' (score and reasoning), 'visual\_blend' (score and reasoning), 'non\_intrusiveness' (score and reasoning), and 'nps\_score' (the average of the three sub-scores).

\textbf{Important Guidelines:}

- Evaluate each dimension independently based on the specific criteria
- Provide clear, specific reasoning for each sub-score
- The final NPS score must be the arithmetic mean of the three sub-scores
- Be objective and consistent in your evaluation

Now, analyze the video and provide your naturalness evaluation.
\end{graybox}

This multi-dimensional evaluation approach provides nuanced assessment of integration quality, capturing the subtle balance between brand visibility and natural scene coherence that is essential for seamless brand integration.

\subsection{LLMScore Implementation}
\label{app:llm_eval}

LLMScore provides a holistic assessment of semantic consistency between the user's original prompt and the generated video by leveraging the reasoning capabilities of multimodal large language models. Unlike VQAScore which decomposes evaluation into multiple discrete questions, LLMScore takes a unified approach where the LLM directly evaluates overall semantic similarity after comprehensively analyzing the video content. 
The LLM evaluation prompt template is shown below:

\begin{graybox}{LLMScore Evaluation Prompt}
\small
You are an expert video content evaluator. Your task is to evaluate the semantic similarity between a text description and a video by comprehensively analyzing how well the video content matches the text semantically.

\textbf{Text Description:}

\textcolor{blue}{\{user\_prompt\}}

\textbf{Evaluation Task:}

Watch the provided video carefully and evaluate how well the video content matches the text description across multiple dimensions.

\textbf{Evaluation Criteria:}

Consider the following aspects when evaluating:

1. \textbf{Content Match:} Does the video show the scenes, objects, actions, and elements described in the text?

2. \textbf{Semantic Consistency:} Does the overall meaning and theme of the video align with the text?

3. \textbf{Details Accuracy:} Are the specific details mentioned in the text reflected in the video?

4. \textbf{Contextual Fit:} Does the video context and setting match what the text describes?

\textbf{Scoring Guidelines:}

- \textbf{1.0:} Perfect match - the video fully and accurately represents everything in the text

- \textbf{0.8-0.9:} Excellent match - the video captures almost all key elements with minor differences

- \textbf{0.6-0.7:} Good match - the video captures most key elements but misses some details

- \textbf{0.4-0.5:} Moderate match - the video partially matches the text but has significant differences

- \textbf{0.2-0.3:} Poor match - the video barely matches the text with major discrepancies

- \textbf{0.0-0.1:} No match - the video content is completely different from the text

\textbf{Output Format:}

Provide your evaluation in JSON format with two fields: 'llm\_score' (a float between 0.0 and 1.0) and 'reasoning' (detailed explanation of your scoring, explaining what matches and what doesn't).

\textbf{Important Guidelines:}

- Be objective and specific in your evaluation
- Consider both what is present and what is missing
- The score should reflect the overall semantic similarity
- Provide clear reasoning that supports your score

Now, evaluate the video and provide your assessment.
\end{graybox}

This comprehensive evaluation approach complements VQAScore by providing a holistic perspective on semantic preservation, capturing subtle semantic relationships that may not be fully captured through discrete question-answering.

\subsection{Human Evaluation Interface and Questionnaire}
\label{app:questionnaire}

To complement automated metrics with subjective user assessments, we conducted a comprehensive human evaluation study using a custom-designed web interface. The evaluation interface was designed to be intuitive and comprehensive, allowing participants to thoroughly assess generated videos across multiple quality dimensions.

\paragraph{Interface Design:}

\begin{figure*}[htbp]
    \centering
    \includegraphics[width=\textwidth]{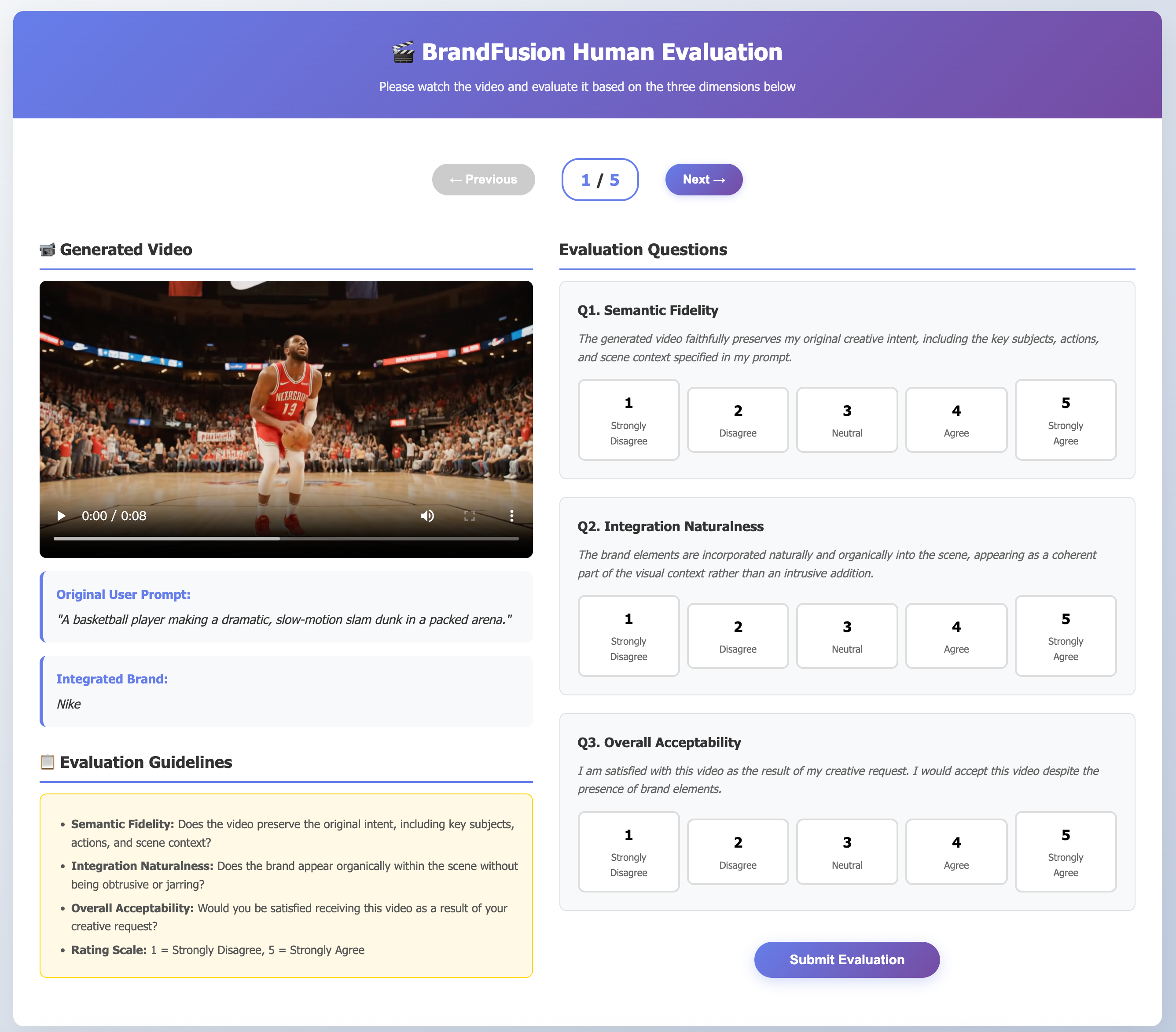}
    \caption{Human evaluation interface showing the video display, contextual information (original prompt and integrated brand), evaluation guidelines, and three assessment questions with 5-point Likert scales.}
    \label{fig:human_eval_interface}
\end{figure*}

Figure~\ref{fig:human_eval_interface} shows our human evaluation interface. The interface is organized into three main sections: (1) {Video Display Panel (Left)}; (2) {Evaluation Questions Panel (Right)}; (3) {Navigation Controls (Top):} 

\paragraph{Evaluation Questionnaire:}
The questionnaire comprises three carefully designed questions that align with our core evaluation objectives:

\textbf{Q1. Semantic Fidelity:}
\textit{"The generated video faithfully preserves my original creative intent, including the key subjects, actions, and scene context specified in my prompt."}
This question assesses whether the brand integration process compromises the user's original creative vision. Higher scores indicate better preservation of semantic content.

\textbf{Q2. Integration Naturalness:}
\textit{"The brand elements are incorporated naturally and organically into the scene, appearing as a coherent part of the visual context rather than an intrusive addition."}
This question evaluates the perceived naturalness and organic quality of brand placement. Higher scores indicate that the brand feels like a natural part of the scene rather than forced product placement.

\textbf{Q3. Overall Acceptability:}
\textit{"I am satisfied with this video as the result of my creative request. I would accept this video despite the presence of brand elements."}
This question captures overall user satisfaction and willingness to accept the video output. It serves as a holistic measure that integrates both semantic quality and integration naturalness from the user's perspective.

\paragraph{Rating Scale:}
All questions use a consistent 5-point Likert scale:
\begin{itemize}
    \item 1 = Strongly Disagree
    \item 2 = Disagree
    \item 3 = Neutral
    \item 4 = Agree
    \item 5 = Strongly Agree
\end{itemize}

This interface design and questionnaire structure enables systematic, reproducible human evaluation while minimizing cognitive load on participants and ensuring consistent interpretation of evaluation criteria.

\section{Additional Experimental Results}
\label{app:additional_exp}

\subsection{Ablation Studies}
\label{app:ablation}

To validate the contribution of each component in BrandFusion's multi-agent framework, we conduct comprehensive ablation studies by systematically removing key agents and analyzing the impact on integration quality. All experiments are conducted on the Veo3 model using the same evaluation protocol as the main experiments.

\subsubsection{Experimental Setup}

We evaluate three ablation variants that progressively remove core components of the multi-agent system:

\begin{itemize}
    \item \textbf{w/o Strategy Generation Agent:} This variant removes the Strategy Generation Agent while maintaining the iterative refinement mechanism through the Critic Agent. Without strategic planning, the Prompt Rewriting Agent must directly determine integration approaches without explicit guidance, relying solely on the brand profile and critic feedback for iterative improvement.
    
    \item \textbf{w/o Critic Agent (Single-pass):} This variant removes the Critic Agent and its iterative refinement mechanism, reducing the framework to single-pass prompt rewriting. The system maintains strategic planning through the Strategy Generation Agent but loses the ability to evaluate and iteratively improve prompts through multiple refinement cycles.
    
    \item \textbf{w/o Strategy \& Critic:} This variant removes both the Strategy Generation Agent and Critic Agent simultaneously, representing the most minimal configuration. The framework degenerates to direct single-pass prompt rewriting without strategic planning or quality assessment, maintaining only brand selection and experience learning capabilities.
\end{itemize}

\subsubsection{Quantitative Results}

Table~\ref{tab:ablation_agents} presents quantitative results across all evaluation dimensions. The complete BrandFusion framework achieves the best performance across all metrics, validating the synergistic contribution of all agents. As expected, performance degrades progressively as components are removed, with the most severe degradation observed when both strategic planning and iterative refinement are absent.

\begin{table*}[!ht]
    \centering
    \caption{Ablation study results on agent components evaluated on Veo3.}
    \label{tab:ablation_agents}
    \resizebox{0.85\textwidth}{!}{%
    \begin{tabular}{@{}lcccccc@{}}
        \toprule
        \textbf{Framework Variant} & \textbf{VBench-Quality} & \textbf{CLIPScore} & \textbf{VQAScore} & \textbf{LLMScore} & \textbf{BPR} & \textbf{NS} \\ 
        \midrule
        BrandFusion (Full) & 0.8283 & 0.3274 & 0.9098 & 0.9556 & 0.9474 & 4.70 \\
        \midrule
        w/o Strategy Generation Agent & 0.8281 & 0.3168 & 0.8945 & 0.9478 & 0.9289 & 4.42 \\
        w/o Critic Agent (Single-pass) & 0.8279 & 0.3021 & 0.8834 & 0.9401 & 0.9045 & 4.15 \\
        w/o Strategy \& Critic & 0.8275 & 0.2912 & 0.8756 & 0.9278 & 0.8812 & 3.82 \\
        \bottomrule
    \end{tabular}%
    }
\end{table*}

The results reveal several important insights about the contribution of each component:

\paragraph{Impact of Strategy Generation Agent.}
Removing the Strategy Generation Agent leads to moderate performance decline: Naturalness Score drops by 0.28 points and Brand Presence Rate decreases by 1.85\%. This demonstrates that explicit strategic reasoning contributes significantly to integration quality. Without strategic guidance, the Prompt Rewriting Agent must make integration decisions reactively, resulting in less optimal placement choices.

\paragraph{Impact of Critic Agent and Iterative Refinement.}
Removing the Critic Agent results in more substantial degradation: Naturalness Score drops by 0.55 points and Brand Presence Rate decreases by 4.29\%. This larger impact indicates that iterative refinement is more critical than strategic planning. Without evaluation and correction capabilities, the framework cannot detect and fix semantic inconsistencies or unnatural brand placements.

\paragraph{Synergistic Effect of Combined Components.}
When both agents are removed simultaneously, performance degradation is most severe: Naturalness Score drops by 0.88 points and Brand Presence Rate falls by 6.62\%. Notably, the combined impact exceeds the sum of individual degradations, suggesting synergistic effects where the Strategy Agent's plans become more valuable when the Critic Agent can verify their execution.

\subsection{Effect of Different LLM Backbones}
\label{app:llm_comparison}

To evaluate the robustness and generalizability of our multi-agent framework across different language model capabilities, we conduct experiments with three LLM backbones of varying capacities. This analysis aims to determine whether BrandFusion's effectiveness depends critically on using the most powerful LLMs, or whether the framework's structured multi-agent design enables strong performance even with more affordable models.

\subsubsection{Experimental Setup}

We evaluate all three LLM backbones on the same benchmark using Veo3 as the T2V model. Each LLM is used consistently across all five agents to ensure fair comparison. All other hyperparameters remain identical across experiments. The three LLM backbones differ significantly in model capacity and inference cost:
\begin{itemize}
    \item \textbf{GPT-4o-mini}: A compact model optimized for efficiency with approximately 8× lower inference cost than GPT-5, suitable for cost-sensitive production deployments.
    \item \textbf{GPT-5}: Our default choice, offering strong reasoning capabilities with balanced cost-performance tradeoff.
    \item \textbf{Gemini-2.5-Pro}: A frontier model with enhanced multimodal understanding and reasoning capabilities, representing the upper bound of current LLM performance.
\end{itemize}

\subsubsection{Quantitative Results}

Table~\ref{tab:llm_comparison} presents comprehensive results across all three LLM backbones. The results demonstrate that BrandFusion maintains strong performance across different model capacities while showing clear improvements as backbone capabilities increase. GPT-4o-mini achieves competitive results despite being the most lightweight model, maintaining 96.2\% of GPT-5's Naturalness Score (4.52 vs. 4.70) and 97.5\% of its Brand Presence Rate (0.9234 vs. 0.9474). Notably, Gemini-2.5-Pro achieves substantial improvements over GPT-5 across all metrics, demonstrating that our framework effectively leverages enhanced model capabilities --- Naturalness Score increases  to 4.85, Brand Presence Rate improves  to 0.9645. These results validate that BrandFusion's structured multi-agent design not only maintains robustness with weaker models through explicit role decomposition and iterative refinement, but also scales effectively to harness the superior reasoning capabilities of frontier LLMs for improved integration quality.

\begin{table*}[!ht]
    \centering
    \caption{Performance comparison across different LLM backbones. All experiments use Veo3 as the T2V model.}
    \label{tab:llm_comparison}
    \resizebox{0.7\textwidth}{!}{%
    \begin{tabular}{@{}lcccccc@{}}
        \toprule
        \textbf{LLM Backbone} & \textbf{VBench} & \textbf{CLIPScore} & \textbf{VQAScore} & \textbf{LLMScore} & \textbf{BPR} & \textbf{NS} \\ 
        \midrule
        GPT-4o-mini & 0.8275 & 0.3198 & 0.8967 & 0.9445 & 0.9234 & 4.52 \\
        GPT-5 (Default) & 0.8283 & 0.3274 & 0.9098 & 0.9556 & 0.9474 & 4.70 \\
        Gemini-2.5-Pro & 0.8295 & 0.3356 & 0.9245 & 0.9687 & 0.9645 & 4.85 \\
        \bottomrule
    \end{tabular}%
    }
\end{table*}
\begin{table*}[!ht]
    \centering
    \caption{LLM backbone performance across different prompt-brand match levels.}
    \label{tab:llm_match_analysis}
    \resizebox{0.65\textwidth}{!}{%
    \begin{tabular}{@{}lcc|cc|cc@{}}
        \toprule
        & \multicolumn{2}{c}{\textbf{High Match}} & \multicolumn{2}{c}{\textbf{Medium Match}} & \multicolumn{2}{c}{\textbf{Low Match}} \\ 
        \cmidrule(lr){2-3} \cmidrule(lr){4-5} \cmidrule(lr){6-7}
        \textbf{LLM Backbone} & LLMScore & NS & LLMScore & NS & LLMScore & NS \\ 
        \midrule
        GPT-4o-mini & 0.9656 & 4.76 & 0.9512 & 4.58 & 0.9167 & 4.18 \\
        GPT-5 (Default) & 0.9734 & 4.90 & 0.9601 & 4.78 & 0.9333 & 4.42 \\
        Gemini-2.5-Pro & 0.9812 & 4.98 & 0.9734 & 4.98 & 0.9515 & 4.75 \\
        \bottomrule
    \end{tabular}%
    }
\end{table*}

\begin{table*}[htbp]
    \centering
    \caption{Cost-performance analysis across LLM backbones. Relative inference cost is normalized to GPT-5 as baseline (1.0×). Naturalness Score per dollar measures integration quality efficiency.}
    \label{tab:llm_cost_analysis}
    \resizebox{0.65\textwidth}{!}{%
    \begin{tabular}{@{}lccc@{}}
        \toprule
        \textbf{LLM Backbone} & \textbf{Relative Cost} & \textbf{Naturalness Score} & \textbf{NS per \$ (Efficiency)} \\ 
        \midrule
        GPT-4o-mini & 0.125× & 4.52 & 36.16 \\
        GPT-5 (Default) & 1.00× & 4.70 & 4.70 \\
        Gemini-2.5-Pro & 1.00× & 4.91 & 4.91 \\
        \bottomrule
    \end{tabular}%
    }
\end{table*}

To better understand how performance scales across different difficulty levels, Table~\ref{tab:llm_match_analysis} presents results stratified by prompt-brand match level. The performance improvements from stronger LLM backbones are most pronounced in challenging Low Match scenarios. Gemini-2.5-Pro achieves a Naturalness Score of 4.75 in Low Match cases, representing a 0.33-point improvement over GPT-5 (4.42) and a 0.57-point gain over GPT-4o-mini (4.18). This pattern demonstrates that our framework effectively translates enhanced reasoning capabilities into improved integration quality, particularly when scene-brand compatibility is inherently challenging and creative strategies are required.

\subsubsection{Cost-Performance Tradeoff}
\label{app:cost}

Table~\ref{tab:llm_cost_analysis} presents a comprehensive cost-performance analysis. GPT-4o-mini offers substantial cost savings (approximately 8× lower than GPT-5) with modest performance degradation, making it suitable for large-scale deployments where cost efficiency is prioritized. GPT-4o-mini achieves the highest "Naturalness Score per dollar" metric (36.16), offering 7.7× better efficiency than GPT-5, making it highly attractive for budget-constrained deployments. Notably, Gemini-2.5-Pro provides meaningful quality improvements over GPT-5 at equivalent inference cost, making it an attractive choice when absolute integration quality is paramount.

In summary, BrandFusion demonstrates strong robustness across LLM backbones, achieving competitive performance even with the cost-efficient GPT-4o-mini while enabling marginal quality gains with premium models like Gemini-2.5-Pro. This robustness validates the effectiveness of our multi-agent framework design, which provides sufficient structure and guidance to enable satisfactory brand integration regardless of underlying LLM capacity. 

\section{Additional Case Studies}
\label{app:cases}

\subsection{Brand Integration Examples}
\label{app:success_cases}

To provide concrete illustrations of BrandFusion's integration capabilities across different brands and difficulty levels, we present detailed case studies for BMW and IKEA. These examples demonstrate how our framework adaptively generates integration strategies based on prompt-brand compatibility, achieving natural placement while maintaining semantic fidelity.

\subsubsection{BMW Brand Integration Across Match Levels}

Figure~\ref{fig:bmw_examples} presents three BMW integration scenarios spanning High, Medium, and Low Match difficulty levels. These cases illustrate how BrandFusion employs distinct strategic approaches based on scene-brand compatibility:

\begin{figure*}[!htbp]
    \centering
    \includegraphics[width=\linewidth]{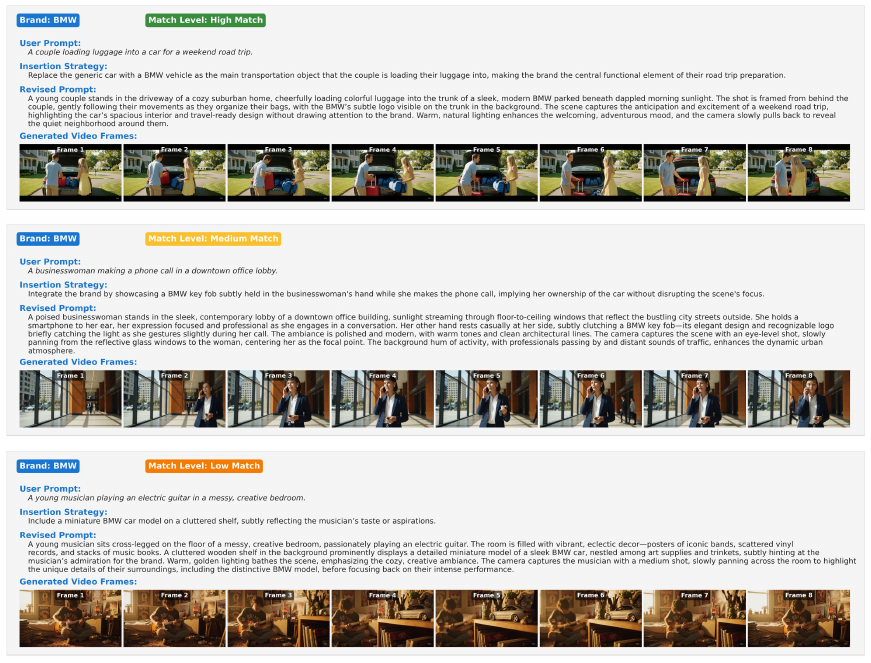}
    \caption{\textbf{BMW brand integration examples across different match levels.} (Top) High Match: BMW vehicle serves as the central functional element for a road trip scenario. (Middle) Medium Match: BMW key fob subtly indicates brand ownership in an office setting. (Bottom) Low Match: Miniature BMW model on shelf reflects the musician's taste and aspirations in a creative bedroom scene.}
    \label{fig:bmw_examples}
\end{figure*}

\paragraph{High Match Scenario: Road Trip Preparation.}
In this naturally compatible scenario, the user prompt describes "a couple loading luggage into a car for a weekend road trip." BrandFusion employs a \textit{Main Object Integration} strategy, replacing the generic vehicle with a BMW car as the primary transportation element. The generated video shows the couple loading colorful luggage into a sleek, modern BMW parked in their driveway, with the brand naturally positioned as the central functional object. The BMW logo is visible on the vehicle's exterior, achieving clear brand presence without disrupting the scene's narrative focus on travel preparation. This represents an ideal integration case where brand and context align seamlessly.

\paragraph{Medium Match Scenario: Office Environment.}
For the prompt "a businesswoman making a phone call in a downtown office lobby," BrandFusion faces moderate integration difficulty as the scene lacks explicit automotive context. The framework adopts a \textit{Subtle Brand Indication} strategy, integrating BMW through a key fob held in the businesswoman's hand. The generated video depicts a poised professional in a modern office lobby, briefly glimpsing a BMW key fob as she gestures during the phone call. This approach implies brand ownership without making the vehicle the scene's focus, maintaining the original emphasis on the business environment while achieving recognizable brand presence through a contextually plausible object.

\paragraph{Low Match Scenario: Creative Bedroom.}
The most challenging case involves "a young musician playing an electric guitar in a messy, creative bedroom"—a scenario with minimal inherent connection to automotive brands. BrandFusion employs a \textit{Lifestyle Representation} strategy, integrating BMW through a detailed miniature car model displayed on a cluttered shelf. The generated video captures the musician's creative space with artistic decorations, music equipment, and personal items, among which the BMW model appears as a subtle reflection of taste and aspirations. The brand presence is understated yet identifiable, avoiding disruption to the scene's artistic focus while successfully incorporating the brand as a natural element of the room's decor.

\subsubsection{IKEA Brand Integration Across Match Levels}

Figure~\ref{fig:ikea_examples} showcases three IKEA integration scenarios demonstrating the framework's versatility with home furnishing brands. These cases highlight different placement strategies ranging from functional product usage to ambient brand presence:

\begin{figure*}[!htbp]
    \centering
    \includegraphics[width=\linewidth]{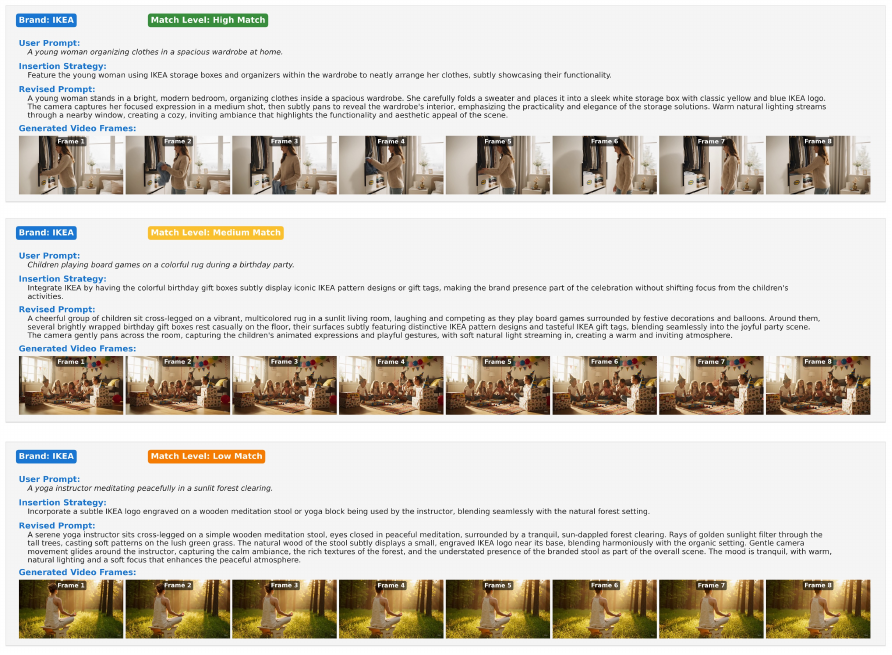}
    \caption{\textbf{IKEA brand integration examples across different match levels.} (Top) High Match: IKEA storage solutions serve as functional organizational tools in a home wardrobe scene. (Middle) Medium Match: IKEA patterns appear on birthday party gift boxes, integrating the brand into a celebratory context. (Bottom) Low Match: IKEA logo on a wooden meditation stool blends harmoniously with a natural forest setting.}
    \label{fig:ikea_examples}
\end{figure*}

\paragraph{High Match Scenario: Home Organization.}
For the prompt "a young woman organizing clothes in a spacious wardrobe at home," BrandFusion identifies strong semantic compatibility and employs a \textit{Functional Product Integration} strategy. The generated video features the woman using distinctive IKEA storage boxes with the classic yellow and blue logo to organize her wardrobe. The boxes are shown being filled with folded sweaters and placed on shelves, naturally showcasing their functionality. Natural lighting through a nearby window illuminates the IKEA branding, achieving clear visibility while maintaining focus on the organizational activity. This exemplifies ideal brand-context alignment where the product serves its intended purpose within the scene.

\paragraph{Medium Match Scenario: Birthday Celebration.}
The prompt "children playing board games on a colorful rug during a birthday party" presents moderate integration difficulty as it lacks explicit furniture context. BrandFusion adopts an \textit{Environmental Integration} strategy, incorporating IKEA through decorative elements. The generated video shows a vibrant birthday party scene with festive decorations and balloons, where several wrapped gift boxes displaying iconic IKEA patterns rest casually near the children. The IKEA branding appears on the gift wrap's distinctive designs and tags, making the brand present as part of the celebration's aesthetic without disrupting focus on the children's activities. This demonstrates creative contextual placement where brand elements enhance scene authenticity.

\paragraph{Low Match Scenario: Forest Meditation.}
The most challenging scenario involves "a yoga instructor meditating peacefully in a sunlit forest clearing"—far removed from typical home furnishing contexts. BrandFusion employs a \textit{Natural Object Integration} strategy, incorporating a wooden meditation stool with a subtle IKEA logo engraved near its base. The generated video captures the serene forest environment with sunlight filtering through trees, where the natural wood of the stool blends harmoniously with the organic setting. The IKEA logo is understated yet identifiable upon closer inspection, integrated as part of the instructor's mindfulness equipment. This case demonstrates sophisticated integration where the brand appears as a natural component of the scene rather than an intrusive commercial element, maintaining the tranquil atmosphere while achieving brand presence.

These detailed examples complement our quantitative evaluation by illustrating the qualitative sophistication of BrandFusion's multi-agent collaboration, showcasing how strategic planning, iterative refinement, and context-aware reasoning converge to achieve seamless brand integration across varying difficulty levels.

\section{Limitations and Ethical Considerations}
\label{app:limitations}

While BrandFusion demonstrates effective brand integration capabilities, we acknowledge several important limitations and ethical considerations that warrant careful attention in real-world deployment.

\subsection{Technical Limitations}
\label{app:tech_limitations}

\paragraph{Dependency on T2V Model Capabilities.}
BrandFusion's integration quality is fundamentally constrained by the underlying T2V model's generation capabilities. When T2V models struggle with specific scene types (e.g., complex physical interactions, fine-grained object details, or rapid motion), brand integration quality correspondingly degrades. Our framework cannot compensate for inherent T2V model limitations in rendering brand elements with high fidelity or maintaining temporal consistency across video frames.

\paragraph{Challenging Integration Scenarios.}
Certain prompt-brand combinations remain fundamentally difficult despite our multi-agent approach. Extremely Low Match scenarios—where semantic compatibility between user intent and brand context is minimal—sometimes result in forced or unnatural integrations that fail to satisfy both semantic preservation and brand visibility requirements. Additionally, prompts involving abstract concepts, purely natural environments without human presence, or historical settings often provide limited natural placement opportunities.

\paragraph{Multi-Brand Integration Complexity.}
The current framework focuses on single-brand integration per video generation request. Extending to simultaneous multi-brand scenarios introduces substantial complexity: brands may compete for visual attention, strategic planning must balance multiple advertiser interests, and evaluation criteria become multidimensional. While conceptually feasible, multi-brand integration requires additional framework development to ensure fair treatment and natural coexistence of multiple commercial elements.

\paragraph{Cultural and Contextual Sensitivity.}
BrandFusion's integration strategies are developed primarily on Western commercial contexts and may not generalize optimally to diverse cultural settings where brand perception, advertising norms, and visual aesthetics differ significantly. Brand placement approaches considered natural in one cultural context might appear inappropriate or ineffective in another. Adapting the framework to respect cultural nuances requires expanding the Brand Knowledge Base with region-specific integration guidelines and cultural sensitivity considerations.

\subsection{Ethical Implications}
\label{app:ethics}

\paragraph{User Consent and Transparency.}
The most critical ethical consideration concerns user awareness and consent regarding brand integration. Users submitting prompts for video generation have a right to know whether and how commercial brands will be incorporated into their content. We strongly advocate for transparent disclosure mechanisms where users are explicitly informed that generated videos may contain brand placements, with options to opt out or select preferred brands. The ecosystem model presented in Figure~\ref{fig:ecosystem} assumes user awareness, but real-world implementations must ensure this assumption holds through clear interface design and user agreement protocols.

\paragraph{Manipulation and Authenticity Concerns.}
Seamless brand integration, by design, aims to make advertisements appear natural and unobtrusive. This raises concerns about subliminal influence and the blurring of boundaries between user-generated creative content and commercial messaging. When brand elements are integrated so naturally that users cannot distinguish them from organic scene components, there is potential for manipulative advertising that circumvents critical engagement. Service providers must balance commercial interests with ethical obligations to preserve user autonomy and prevent deceptive practices.

\paragraph{Impact on Creative Expression.}
While BrandFusion prioritizes semantic fidelity to user prompts, brand integration inevitably constrains creative freedom to some degree. Users may feel that commercial elements compromise their artistic vision or dilute their intended message. This tension between monetization and creative autonomy is inherent to ad-supported content models. We recommend providing users with control mechanisms—such as brand category preferences, integration intensity settings, or premium ad-free tiers—to mitigate this concern and respect individual creative priorities.

\paragraph{Equity and Access Considerations.}
Ad-supported T2V services enabled by frameworks like BrandFusion could democratize access to expensive video generation technologies by offsetting computational costs. However, this model may also create inequities where users unable to afford premium ad-free options receive lower-quality experiences with intrusive brand placements. Service providers should carefully balance revenue generation with equitable access, potentially offering free tiers with reasonable integration constraints rather than exploitative advertising loads.

\paragraph{Data Privacy and Brand Profiling.}
To optimize brand selection and integration strategies, systems may collect user preference data, prompt histories, and interaction patterns. This data collection raises privacy concerns regarding user profiling for targeted advertising. Service providers must implement robust data protection measures, clearly communicate data usage policies, and provide users with data control and deletion rights in compliance with privacy regulations (e.g., GDPR, CCPA).

\subsection{Potential Misuse and Safeguards}
\label{app:safeguards}

\paragraph{Unauthorized Brand Usage.}
BrandFusion could potentially be misused to generate videos containing brands without proper authorization or licensing agreements. Malicious actors might integrate competitor brands to create false associations, generate counterfeit content, or produce misleading advertisements. To prevent unauthorized usage, we recommend implementing strict brand verification mechanisms where only registered brands with verified ownership can be integrated. Service providers should maintain comprehensive Brand Knowledge Bases with authenticated brand profiles and legal authorization records.

\paragraph{Inappropriate Content Associations.}
Without proper safeguards, brands could be integrated into inappropriate, offensive, or harmful content that damages brand reputation or violates advertising standards. For example, alcohol brands appearing in content involving minors, or luxury brands associated with illegal activities. Our framework includes a ``Prohibited Scenarios" field in each brand profile (Section~\ref{app:knowledge_base}) to specify contextual restrictions. We strongly recommend expanding this mechanism with automated content moderation systems that detect potentially inappropriate prompt-brand combinations and reject integration requests that violate ethical guidelines or legal restrictions.

\paragraph{Children's Content Protections.}
Brand integration in content targeting or involving children raises heightened ethical concerns and regulatory compliance issues (e.g., COPPA in the United States, GDPR's child protection provisions in Europe). Children may be particularly susceptible to integrated advertising and less capable of recognizing commercial intent. We advocate for strict policies prohibiting brand integration in content explicitly identified as child-directed, implementing age verification mechanisms, and applying enhanced scrutiny to any integration involving minors or child-related contexts.

\paragraph{Misinformation and Deceptive Content.}
The framework could potentially be misused to create deceptive videos where brands are associated with false claims, fabricated testimonials, or misleading contexts. For instance, generating videos that falsely suggest brand endorsements by public figures or misrepresent product capabilities. Safeguards should include content verification systems, explicit disclaimers indicating AI-generated content, and prohibition of integration in news-style or documentary formats where authenticity expectations are high.

\paragraph{Recommended Safeguard Framework.}
Based on these considerations, we recommend service providers implementing BrandFusion adopt a comprehensive safeguard framework:

\begin{itemize}
    \item \textbf{Transparent Disclosure}: Clear visual indicators (watermarks, labels) identifying AI-generated content with brand integration
    \item \textbf{User Consent Mechanisms}: Explicit opt-in requirements and granular control over brand integration preferences
    \item \textbf{Content Moderation}: Automated and human review systems to detect policy violations, inappropriate associations, and harmful content
    \item \textbf{Brand Authorization Verification}: Strict authentication protocols ensuring only legitimately registered brands are integrated
    \item \textbf{Regulatory Compliance}: Adherence to advertising standards, consumer protection laws, and platform-specific policies
    \item \textbf{Appeal and Redress Mechanisms}: Processes for users and brands to report misuse and request content removal
    \item \textbf{Continuous Monitoring}: Regular audits of generated content and integration patterns to identify emerging misuse trends
\end{itemize}

\paragraph{Research Community Responsibility.}
As researchers introducing brand integration capabilities for T2V generation, we acknowledge responsibility for anticipating and mitigating potential harms. We encourage the research community to engage in ongoing dialogue about ethical AI deployment in advertising contexts, develop shared best practices, and collaborate with policymakers to establish appropriate regulatory frameworks. The goal should be creating sustainable T2V monetization pathways that respect user rights, protect vulnerable populations, and maintain public trust in AI-generated media.

In summary, while BrandFusion offers technical solutions for seamless brand integration, successful real-world deployment requires careful attention to technical limitations, ethical implications, and potential misuse scenarios. Service providers must implement comprehensive safeguards, maintain transparency with users, and prioritize ethical considerations alongside commercial objectives. Only through responsible deployment can brand integration contribute positively to sustainable T2V service ecosystems.

\end{document}